\documentclass[11pt]{article}
\usepackage[margin=1in]{geometry}
\usepackage[T1]{fontenc}
\usepackage{lmodern}
\usepackage{amsmath,amssymb,amsthm}
\usepackage[round]{natbib}
\usepackage{booktabs}
\usepackage{tikz}
\usetikzlibrary{arrows.meta}
\usepackage[expansion=false]{microtype}
\PassOptionsToPackage{hyphens}{url}
\usepackage[colorlinks=true,linkcolor=black,citecolor=black,urlcolor=black]{hyperref}
\hypersetup{pdftitle={The Value of Labels in Data Selection for Least Squares},pdfauthor={Evan Borodow}}
\newtheorem{theorem}{Theorem}
\newtheorem{lemma}[theorem]{Lemma}
\newtheorem{proposition}[theorem]{Proposition}
\newtheorem{corollary}[theorem]{Corollary}
\theoremstyle{definition}
\newtheorem{definition}[theorem]{Definition}

\newcommand{\R}{\mathbb R}
\newcommand{\Z}{\mathbb Z}
\newcommand{\E}{\mathbb E}
\newcommand{\tr}{\operatorname{tr}}
\newcommand{\be}{\mathbf e}
\newcommand{\av}[1]{\langle #1\rangle}
\newcommand{\Prob}{\mathbb P}
\newcommand{\Var}{\operatorname{Var}}
\newcommand{\Pois}{\operatorname{Pois}}
\newcommand{\sgn}{\operatorname{sgn}}
\newcommand{\one}{\mathbf 1}
\newcommand{\cutoff}{{\bar\lambda}}

\title{The Value of Labels in Data Selection\\ for Least Squares}
\author{Evan Borodow}
\date{}

\begin{document}

\maketitle

\begin{abstract}
Suppose a selector sees a finite labeled dataset and chooses $n$ of its rows, with repetition. Ordinary least squares is then fitted to the rows that were chosen, without weights, and the excess loss of the fit on the full dataset is measured in units of the optimal loss. Without further assumptions its worst case does not decay with $n$. We examine the problem for $d$ features, $m$ labels per row and rows of normalized leverage at most $\kappa$, and determine or bound the leading constants. A selector that must establish its rule before seeing the labels has an optimal worst-case ratio of $d\kappa/n$ to leading order, whatever use it makes of the features. With scalar labels, a selector that sees the labels obtains a value between $d\kappa/(2n)$ and $c_*d\kappa/n$ to leading order, where $c_*<1$ does not depend on $d$, so that its gain does not disappear as $d$ grows. For $d\le4$, the constant is $d\kappa/2$, so that labels are worth exactly a factor of two there. The selector rounds the counts of rare rows and, by changing the proportions of the frequent rows, reshapes its own design matrix. With two or more labels an adversary is able to use one label to penalize these changes. The constants then lie between $d\kappa\,\Gamma_{d,m-1}$ and $d\kappa\,\Gamma_{d,m}$, where $\Gamma_{d,p}$ is the constant of rounding sums of rank-one $d\times p$ matrices, and for fixed $m\ge2$ labels are worth a factor that tends to one as $d$ grows. For grouped designs with scalar labels, in which reshaping amounts to the allocation of a common budget, the constants are $27d/128$ up to an additive constant, compared with $d$ without labels. Nevertheless, for scalar labels $d\kappa/2$ is not the constant in every dimension: for every $\kappa>1$, every $c<e^{-1/2}$ and all large $d$ we construct datasets on which every selection has an excess ratio of at least $c\,d\kappa/n$ when $n$ is large.
\end{abstract}


\section{Introduction}\label{sec:intro}

In data selection a learning rule is fixed and an informed selector chooses a small training set from a larger dataset so that the rule, which is trained on the selection, performs well on the whole dataset \citep{hanneke2025erm}. The selection is \emph{unweighted}: the rule sees only the chosen examples. A series of recent works provides a detailed picture of data selection for the empirical mean under the squared loss. With $n$ selected points in $\R^D$ the best worst-case excess loss, expressed in units of the data variance, is $1/(2n-1)$ for $D\le3$ \citep{hanneke2025erm,huang2026low}, and in every fixed dimension it is $\Lambda_D/n$ to leading order, where $\Lambda_D$ is the constant of an integer-rounding problem with $\Lambda_D=\frac12$ for $D\le4$, $\Lambda_D>\frac12$ for $D\ge5$ and $1-\Lambda_D=\Theta(D^{-1/3})$ \citep{dewasurendra2026barrier,qin2026uniform,mishra2026cone}.

For linear regression the results are quite different. \citet{hanneke2025open} posed the question of the worst-case cost of unweighted selection for least squares. The answer is degenerate: for $d$ features the worst-case excess ratio is $d$ for every budget $n\ge d$, so additional rows do not help: see \citet[Theorem~8 and Section~5]{hanneke2025erm} for $n=d$, $m=1$, \citet{peng2026pipeline} in general and, independently, \citet{huang2026regression} for $m=1$. The extremal datasets contain rows of extremely large leverage. \citet[Section~6.4]{huang2026regression} noted that a regularity assumption such as bounded leverage could restore a dependence on the budget. This paper executes that plan, determines or bounds the leading constants, and compares selection with and without access to the labels.

\paragraph{Setting.} Consider a dataset that consists of rows $(x_i,y_i)\in\R^d\times\R^m$ whose second-moment matrix $\Sigma$ is nonsingular; its \emph{normalized leverage} is $\max_ix_i^\top\Sigma^{-1}x_i/d\ge1$. The selector chooses $n$ rows with repetition, and the learner returns the minimum-norm least-squares fit of the chosen rows. The performance is measured as the \emph{excess ratio}, the excess loss of this fit on the full dataset divided by the optimal loss. Let $A^{(m)}_{d,\kappa}(n)$ be the worst case of the best ratio over datasets whose normalized leverage is at most $\kappa$, and let $B^{(m)}_{d,\kappa}(n)$ be the same quantity, with the expected ratio, for a selector that sees the features but must fix a (possibly randomized) rule before the labels are revealed. Precise definitions are in Section~\ref{sec:setting}. Except for Theorem~\ref{thm:group}(i), all theorems about data selection are asymptotic in $n$ with $d$, $m$ and $\kappa$ fixed.

\paragraph{Results.} Let
\[
 \underline\alpha^{(m)}_{d,\kappa}=\liminf_{n\to\infty}\frac{n\,A^{(m)}_{d,\kappa}(n)}{d\kappa},\qquad
 \overline\alpha^{(m)}_{d,\kappa}=\limsup_{n\to\infty}\frac{n\,A^{(m)}_{d,\kappa}(n)}{d\kappa}
\]
be the lower and upper leading constants in units of $d\kappa$; we do not know whether they always agree.
\begin{itemize}
\item \emph{Without labels} (Theorem~\ref{thm:blind}). For all $d$, $m$ and $\kappa$, $\lim_nn\,B^{(m)}_{d,\kappa}(n)=d\kappa$. Uniform sampling, restricted to selections whose design matrix is well conditioned, is asymptotically minimax optimal. In other words, in the worst case over designs of normalized leverage at most $\kappa$, where the design may vary with $n$, no rule that depends on the features only, including rules based on leverage scores or optimal designs, improves the constant.
\item \emph{One label} (Theorem~\ref{thm:one}). There is an absolute constant $c_*<1$ with $\frac12\le\underline\alpha^{(1)}_{d,\kappa}\le\overline\alpha^{(1)}_{d,\kappa}\le\min\{\Lambda_d,c_*\}$ for all $d$ and $\kappa$. Plain rounding gives $\Lambda_d$, which tends to one; the bound $c_*$ does not depend on $d$. For $d\le4$ both constants are $\frac12$. One half is not the answer in every dimension. For every $\kappa>1$ and every $c<e^{-1/2}\approx0.607$, the lower leading constant $\underline\alpha^{(1)}_{d,\kappa}$ is above $c$ for all large $d$, and, still for $\kappa>1$, it is above $\frac12$ already when $d\ge332\,101$ and $d\kappa\ge332\,100\,000$ (Theorem~\ref{thm:barrier}). A randomized algorithm achieves a constant $c<1$ with probability at least $1-\epsilon$, with a number of operations polynomial in $N$, $d$, $n$ and $\log(1/\epsilon)$ in the real-arithmetic model with an oracle for linear programs (Theorem~\ref{thm:algo}); here $N$ is the number of rows.
\item \emph{Several labels} (Theorem~\ref{thm:several}). For all $d\ge1$, $m\ge2$ and $\kappa\ge1$,
\[
 \Gamma_{d,m-1}\ \le\ \underline\alpha^{(m)}_{d,\kappa}\ \le\ \overline\alpha^{(m)}_{d,\kappa}\ \le\ \Gamma_{d,m},
\]
where $\Gamma_{d,p}$ is the rounding constant for rank-one matrices in $\R^{d\times p}$, $\Gamma_{d,1}=\Lambda_d$ and $\max\{\Lambda_d,\Lambda_p\}\le\Gamma_{d,p}\le\Lambda_{dp}$. For a fixed $m\ge2$ and $\kappa$, both constants are $1-\Theta(d^{-1/3})$ as $d\to\infty$. If $dm\le4$ they are $\frac12$.
\item \emph{Value of labels} (Corollary~\ref{cor:value}). For one label, $\liminf_nB/A\ge1/c_*>1$ in every dimension, and $\limsup_d\limsup_nB/A\le\sqrt e$ if $\kappa>1$; however, for fixed $m\ge2$, $\limsup_nB/A\to1$ as $d\to\infty$.
\item \emph{Grouped designs} (Theorem~\ref{thm:group}). When the features are indicators of $d\ge2$ groups and the labels are scalar, the fit consists of $d$ group means and reshaping is the allocation of a common budget. Without a leverage cap, and without the normalization by $d\kappa$, the label-aware constants then lie between $27d/128$ and $27d/128+13/24$, and the label-blind constant is exactly $d$. Here labels are worth a factor that tends to $128/27$ as $d$ grows.
\end{itemize}
Table~\ref{tab:constants} summarizes the constants. The proof gives $c_*=1-10^{-2364}$. The asymptotic formula $1-\Lambda_d\sim(27/(4d))^{1/3}$ suggests that this improves on $\Lambda_d$ only for $d$ above about $7\cdot10^{7092}$.

\begin{table}[t]
\centering
\renewcommand{\arraystretch}{1.15}
\begin{tabular}{@{}lll@{}}
\toprule
Case & $\underline\alpha^{(m)}_{d,\kappa}$ and $\overline\alpha^{(m)}_{d,\kappa}$ & Theorem \\
\midrule
$d=1$, any $m$ & $\Lambda_m$ & \ref{thm:one}, \ref{thm:several} \\
$dm\le4$ & $\frac12$ & \ref{thm:one}, \ref{thm:several}\\
$m=1$, $d\ge5$ & in $[\frac12,\,\min\{\Lambda_d,c_*\}]$, with $c_*<1$ absolute & \ref{thm:one}\\
$m=1$, $\kappa>1$, $d\to\infty$ & in $[e^{-1/2}-o(1),\,c_*]$ & \ref{thm:one}, \ref{thm:barrier}\\
$m\ge2$ & in $[\Gamma_{d,m-1},\,\Gamma_{d,m}]$ & \ref{thm:several}\\
$m\ge2$ fixed, $d\to\infty$ & $1-\Theta(d^{-1/3})$ & \ref{thm:several}\\
\bottomrule
\end{tabular}
\caption{Lower and upper leading constants of label-aware selection, in units of $d\kappa$. The label-blind constant is $1$ in every case (Theorem~\ref{thm:blind}). Here $n\to\infty$ before $d\to\infty$, with $\kappa$ fixed. An interval contains both constants and does not assert that they agree.}
\label{tab:constants}
\end{table}

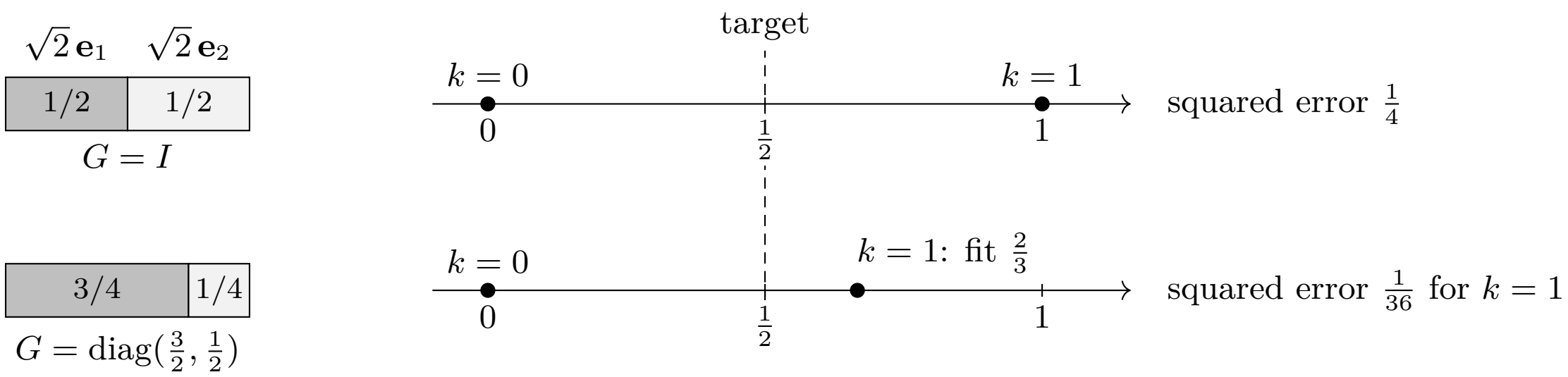
\begin{figure}[t]
\centering
\begin{tikzpicture}[x=5.2cm,y=1cm,font=\footnotesize]
\draw[densely dashed] (0.5,-0.1)--(0.5,2.25);
\node[anchor=south] at (-0.76,2.02) {$\sqrt2\,\be_1$};
\node[anchor=south] at (-0.54,2.02) {$\sqrt2\,\be_2$};
\foreach \y/\s/\p/\q/\g in {1.75/-0.65/{1/2}/{1/2}/{G=I},0/-0.54/{3/4}/{1/4}/{G=\mathrm{diag}(\frac32,\frac12)}}{
 \draw[fill=black!25] (-0.87,\y-0.25) rectangle (\s,\y+0.25);
 \draw[fill=black!5] (\s,\y-0.25) rectangle (-0.43,\y+0.25);
 \node[font=\scriptsize] at ({(-0.87+\s)/2},\y) {$\p$};
 \node[font=\scriptsize] at ({(\s-0.43)/2},\y) {$\q$};
 \node[anchor=north] at (-0.65,\y-0.25) {$\g$};
 \draw[->] (-0.1,\y)--(1.16,\y);
 \foreach \t/\l in {0/0,0.5/{\frac12},1/1} {\draw (\t,\y-0.06)--(\t,\y+0.06); \node[anchor=north,fill=white,inner sep=1.5pt] at (\t,\y-0.09) {$\l$};}
}
\node[anchor=south] at (0.5,2.22) {target};
\fill (0,1.75) circle (2pt) (1,1.75) circle (2pt) (0.6667,0) circle (2pt);
\fill (0,0) circle (2pt);
\node[anchor=south] at (0,1.78) {$k=0$};
\node[anchor=south] at (1,1.78) {$k=1$};
\node[anchor=south] at (0,0.03) {$k=0$};
\node[anchor=south west] at (0.64,0.03) {$k=1$: fit $\frac23$};
\node[anchor=west] at (1.2,1.75) {squared error $\frac14$};
\node[anchor=west] at (1.2,0) {squared error $\frac1{36}$ for $k=1$};
\end{tikzpicture}
\caption{Design reshaping. Left: selected proportions of the two frequent types and the resulting design $G$. Right: first coordinate of the fit when the rare row, with gradient $\be_1$ and coefficient $\frac12$, is selected $k$ times; the target is $\frac12$. With the population proportions (top) both choices $k=0,1$ have absolute error $\frac12$, the rounding error. With proportions $\frac34,\frac14$ (bottom) selecting the rare row once moves the fit to $1/\frac32=\frac23$, with absolute error $\frac16$.}
\label{fig:reshape}
\end{figure}

\paragraph{Rounding and reshaping.} The selector may choose the ordinary, or \emph{frequent}, rows in proportions different from their population proportions. This changes the selected design matrix, and therefore also the map from the selected gradient to the fitted parameter. The following example uses the game of Section~\ref{sec:game}, in which the selected design is divided by the budget and fits are scaled by $\sqrt n$. Let the frequent types be $\sqrt2\,\be_1$ and $\sqrt2\,\be_2$ with equal masses and zero residuals with respect to their own fit, and let there be one \emph{rare} row, that is, a row with a large residual, with gradient $g=\be_1$ and coefficient $a=\frac12$ (its count in a proportional selection); the target, the fit of the full dataset, is $r=ag=\frac12\be_1$, and the rare energy $a\|g\|^2$ is $\frac12$. At the population proportions the design is $I$, and selecting the rare row once or not at all leaves the squared error $\frac14$, one half of the energy. Selecting it once and the frequent types in proportions $\frac34,\frac14$ gives the design $\operatorname{diag}(\frac32,\frac12)$, the fit $\frac23\be_1$ and the squared error $(\frac23-\frac12)^2=\frac1{36}$ (Figure~\ref{fig:reshape}). The label-aware problem is therefore not a rounding problem: the selector rounds the counts of the few rows with large residuals and, at the same time, reshapes the design.

\paragraph{Why one label is different.} An adversary can try to prevent reshaping by giving the frequent rows residuals, so that a change of their proportions moves the fit. With $m\ge2$ labels this works. In our construction the frequent rows form nearly a regular simplex and their last label is nearly constant, and every selection with bounded $n\rho$ must then use them nearly in their population proportions: the last label \emph{freezes the design}. Rounding in the other $m-1$ labels remains, and the constants are at least $\Gamma_{d,m-1}$. With one label, the residuals that penalize reshaping lie in the label that has to be corrected, so that reshaping can also be used to correct it. If neither random counts nor the small changes of the proportions used in the proof gain, then almost all of the rare energy comes from pairs for which the effect of the rare row nearly cancels the residual of the frequent row. Selecting the rare row twice overshoots, and then a change of the proportions by a constant factor gains (Section~\ref{sec:one}). This gives $c_*$. The gain is limited: if $\kappa>1$ and $d$ is large, the adversary can use rare gradients with coordinates whose sign no change of the proportions can reverse, which gives $e^{-1/2}$ (Section~\ref{sec:barrier}).

\paragraph{Related work.} The data-selection framework and the mean and regression questions were introduced by \citet{hanneke2025erm,hanneke2025open}. For the mean, exact values and the leading constant were obtained by, among others, \citet{huang2026low}, \citet{dewasurendra2026barrier}, \citet{qin2026uniform}, \citet{zheng2026four} and \citet{mishra2026cone}. For regression, the unweighted problem without regularity assumptions was resolved by \citet{peng2026pipeline} and, independently, for $m=1$ by \citet{huang2026regression}, and weighted selection, in which the learner receives weights with the rows and which was introduced alongside the regression problem by \citet{hanneke2025erm}, was examined further by, among others, \citet{huang2026regression}, \citet{zhang2026vector,zhang2026weighted} and \citet{liu2026weighted}. To our knowledge none of these works contains a rate in the budget for unweighted, unregularized least squares under bounded leverage beyond the mean problem. \citet[Propositions~8.1 and~8.2]{mishra2026cone} treats smooth objectives whose selected versions are uniformly strongly convex, which covers ridge regression, and objectives that share one quadratic part, which corresponds to our case \mbox{$\kappa=1$, $d=1$}.

Label-blind selection of rows for least squares is the subject of experimental design \citep{pukelsheim2006}, of leverage-score sampling \citep{ma2015}, of volume sampling \citep{derezinski2017,derezinski2018,derezinski2019} and of linear-sample sparsification \citep{chen2019}. The majority of these methods reweight the selected rows. Unweighted fits are used, for instance, with exact designs, in one of the estimators of \citet{ma2015}, and in volume sampling \citep{derezinski2017,derezinski2018reverse}, whose unweighted fit is unbiased but can be poor at high accuracy \citep{derezinski2018}. \citet{derezinski2019} define a minimax value for random label-blind designs with unbiased estimators and arbitrary responses and bound it from above without dependence on the leverage, with a lower bound that matches within a constant factor for some designs. The upper bound of Theorem~\ref{thm:blind} follows the classical method for independent sampling under bounded statistical leverage \citep{hsu2014,mourtada2022}. The matching lower bound against every label-blind rule appears to be new. For one constant feature it corresponds to classical minimax results for the sample mean under simple random sampling \citep{aggarwal1959,bickel1981}. \citet{kolossov2024} analyze selection without reweighting for independent data by rules that do not see the labels (they may use a surrogate model), and observe that the optimal rule changes the covariance of the selected features. Among methods that use the responses, \citet{boutsidis2013} construct reweighted coresets that depend on the labels and prove that with $N+d$ labels per row no selection of $n>d$ rows with weights, even one that sees the labels, has a worst-case excess ratio below about $d/n$ for large $N$; here the number of labels is fixed, and leading constants are compared. \citet{sabato2014} compare passive sampling with label-informed weighted sampling for misspecified linear regression, optimal subsampling \citep[for example,][for logistic regression]{wang2018} samples rows independently and reweights them, and \citet{wang2019more} replaces the weights by a bias correction. Independent uniform draws do not coordinate the counts so as to create cancellations between the gradients (feature times residual) of the selected rows; this coordination is the source of the factor one half for $dm\le4$.

\paragraph{Organization.} Sections~\ref{sec:setting} and~\ref{sec:blind} give the setting and the label-blind result, Sections~\ref{sec:game} to~\ref{sec:barrier} treat one label, Section~\ref{sec:several} several labels and the value of labels, Section~\ref{sec:group} grouped designs and Section~\ref{sec:disc} open questions. Appendix~\ref{app:notation} is a table of notation.

\section{Setting and rounding constants}\label{sec:setting}

Throughout, $n,m,d\ge1$ are integers and $\kappa\ge1$ is real.

\paragraph{Datasets and selections.} A dataset $D$ is a finite indexed family of rows $(x_i,y_i)\in\R^d\times\R^m$, $i\in[N]$, with $\Sigma_D=N^{-1}\sum_ix_ix_i^\top$ nonsingular. The \emph{mass} of a row is its fraction of the $N$ rows, counted with repetitions; positive rational masses that sum to one determine a dataset up to repetition of all rows, which preserves the attainable fits and losses on this fixed labeled dataset. For $W\in\R^{m\times d}$ let $L_D(W)=N^{-1}\sum_i\|y_i-Wx_i\|^2$, let $W^*$ be its minimizer and $L_D^*=L_D(W^*)$. The \emph{normalized leverage} of $D$ is $\kappa_D=\max_ix_i^\top\Sigma_D^{-1}x_i/d\ge1$, where the inequality holds because the average of $x_i^\top\Sigma_D^{-1}x_i$ is $d$. A \emph{selection} of budget $n$ is a count vector $k\in\Z_{\ge0}^N$ with $\sum_ik_i=n$, and $\hat W_k$ is the minimizer of $\sum_ik_i\|y_i-Wx_i\|^2$ of minimum Frobenius norm. The \emph{excess ratio} is $\rho_D(k)=(L_D(\hat W_k)-L_D^*)/L_D^*$, defined when $L_D^*>0$.

\begin{definition}\label{def:AB}
Let $A^{(m)}_{d,\kappa}(n)=\sup_{D}\min_{k}\rho_D(k)$, where the supremum is over datasets in $\R^d\times\R^m$ with $\kappa_D\le\kappa$ and $L_D^*>0$. For the label-blind problem, a \emph{design} is a family $X=(x_i)_{i\in[N]}$ with $N>d$ and nonsingular second moment, and
\[
 B^{(m)}_{d,\kappa}(n)=\sup_{X}\ \inf_{Q}\ \sup_{Y}\ \E_{k\sim Q}\,\rho_{(X,Y)}(k),
\]
where $X$ ranges over designs with normalized leverage at most $\kappa$, $Q$ over probability laws on selections of budget $n$ (which may depend on $X$), and $Y$ over label families with $L^*_{(X,Y)}>0$.
\end{definition}

We abbreviate $A_{d,\kappa}=A^{(1)}_{d,\kappa}$ and $B_{d,\kappa}=B^{(1)}_{d,\kappa}$. Clearly $B^{(m)}_{d,\kappa}(n)\ge A^{(m)}_{d,\kappa}(n)$. If the \emph{selected design} $G_k=\sum_ik_ix_ix_i^\top$ is nonsingular and $e_i=y_i-W^*x_i$ are the residuals of the full fit, then
\begin{equation}\label{eq:defect}
 (\hat W_k-W^*)^\top=G_k^{-1}\sum_ik_ix_ie_i^\top ,
\end{equation}
the normal-equation form of the defect identity of \citet[Proposition~2.1]{huang2026regression}. Since $\sum_ix_ie_i^\top=0$, the selector must make the selected \emph{gradient} $\sum_ik_ix_ie_i^\top$ small relative to $G_k$. For nonsingular $G_k$ the ratio $\rho_D(k)$ is unchanged when every feature $x_i$ is replaced by $Tx_i$ with $T$ invertible, or every label $y_i$ by $y_i+W_0x_i$, and so is $\kappa_D$. In upper bounds, where the constructed selections have nonsingular designs, we may therefore \emph{whiten}, that is, assume $\Sigma_D=I$; then $\|x_i\|^2\le d\kappa$ and the excess loss is $\|\hat W_k-W^*\|_F^2$. For singular $G_k$ the minimum-norm fit is not equivariant, and we never transform such selections.

\paragraph{Rounding constants.} Let $D,d,p\ge1$ be integers. For $g_1,\dots,g_q$ in a Euclidean space and $a\in[0,\infty)^q$ with energy $C(g,a)=\sum_ia_i\|g_i\|^2>0$ let
\begin{equation*}
 \Lambda(g,a)=\inf_{k\in\Z_{\ge0}^q}\frac{\|\sum_i(k_i-a_i)g_i\|^2}{C(g,a)},\qquad
 \Lambda_D=\sup\Lambda(g,a),\qquad
 \Gamma_{d,p}=\sup_{\operatorname{rank}g_i\le1}\Lambda(g,a),
\end{equation*}
where the first supremum is over all $q$, $g\in(\R^D)^q$ and $a$, and the second over all $q$, $a$ and $g\in(\R^{d\times p})^q$ with $\operatorname{rank}g_i\le1$, with Frobenius norms. The constant $\Lambda_D$ is the leading constant of mean selection in $\R^D$: the configuration $(g,a)$ stands, for rational $a_i$ and large $n$, for a dataset with a fraction $a_i/n$ of rare points at $\sqrt n\,g_i$, and $k_i$ is the number of selected copies. The gradients $x_ie_i^\top$ of least squares have rank at most one, which is why $\Gamma$ appears.

\begin{proposition}\label{prop:Lambda}
(i) $\frac12\le\Lambda_D\le\Lambda_{D+1}\le1$. In the definitions of $\Lambda_D$ and $\Gamma_{d,p}$ one may restrict to linearly independent families with rational $a_i\in(0,1)$. (ii) $\Lambda_D=\frac12$ for $D\le4$. (iii) $\Gamma_{d,1}=\Lambda_d$, $\Gamma_{d,p}=\Gamma_{p,d}$, and $\max\{\Lambda_d,\Lambda_p\}\le\Gamma_{d,p}\le\Lambda_{dp}$. Moreover $\Gamma_{d,p}$ is nondecreasing in each index. (iv) $\Gamma_{d,p}=\frac12$ if and only if $dp\le4$.
\end{proposition}

Part~(ii) is contained in \citet{qin2026uniform} and \citet{mishra2026cone}; Appendix~\ref{app:round} contains a short proof of the proposition. Moreover $\Lambda_D<1$ for every $D$ \citep[Theorem~7.1]{mishra2026cone}, $\Lambda_5=(6-2\sqrt3)/5\approx0.507$ \citep{mishra2026cone,qin2026uniform}, and $1-\Lambda_D\sim(27/(4D))^{1/3}$ \citep[Theorem~2.3]{qin2026uniform}. The asymptotic formula is used only for the statements with $d\to\infty$ and for the comparison of $c_*$ with $\Lambda_d$ in Section~\ref{sec:intro}.

\section{Selection without labels}\label{sec:blind}

\begin{theorem}[Label-blind selection]\label{thm:blind}
For all $d,m\ge1$ and $\kappa\ge1$, $\lim_{n\to\infty}n\,B^{(m)}_{d,\kappa}(n)=d\kappa$.
\end{theorem}

The upper bound is attained by uniform sampling, conditioned on the event that the selected design is well conditioned. For the lower bound the design consists of a cluster of identical rows $\sqrt{d\kappa}\,\be_1$ and of frequent rows whose first coordinate is constant. One label family punishes every deviation of the share of the cluster from its proportional share of the budget; for a proportional share, random signs on the cluster cost $d\kappa/n$. The proof is in Appendix~\ref{app:blind}; it gives $n\,B^{(m)}_{d,\kappa}(n)\le d\kappa\,(1+O(\sqrt{\log n/n}))$ for fixed $d$ and $\kappa$.

The worst datasets must vary with the budget. For a fixed design with $N$ rows, the rounded uniform counts with $|k_i-n/N|\le1$ have $\rho\le N^2/(n-N)^2$ for all labels with $L^*>0$ and all $n>N$: after whitening, $G_k\succeq(n-N)I$, and since $\sum_ix_ie_i^\top=0$ the selected gradient is $\sum_i(k_i-n/N)x_ie_i^\top$, with squared norm at most $N\cdot N L^*$, because after whitening the matrix with rows $x_i^\top$ has operator norm $\sqrt N$. Thus for a fixed dataset the ratio is $O(n^{-2})$ even for a selector that does not see the labels; in our lower bounds the rate $1/n$ comes from individual rows whose expected count under proportional selection stays bounded.

\section{The reshaping game}\label{sec:game}

In this section and the next two the labels are scalar. After whitening, scaling to $L^*=1$, and a reduction of the support by a linear program, call a row \emph{rare} if its residual exceeds $\eta\sqrt n$ in absolute value, where $\eta>0$ is small and fixed, and \emph{frequent} otherwise. A rare row of mass $a_i/n$ has the scaled gradient $g_i=x_ie_i/\sqrt n$, where $e_i$ is its residual. The following game describes what the selector can achieve; Theorem~\ref{thm:data} in Appendix~\ref{app:one} transfers the bound of Theorem~\ref{thm:U} to datasets. The game is used for upper bounds only. All lower bounds are proved directly for datasets.

An \emph{affine frame} in $\R^d$ is a finite probability space of \emph{types} $x\in\R^d$ with positive masses $\nu$ and a function $\phi$ on the types with $\E_\nu xx^\top=I$ and $\E_\nu[\phi x]=0$; repeated vectors with different values of $\phi$ are allowed. It stands for the frequent rows; $\phi$, of arbitrary size, is their scaled residual with respect to their own least-squares fit, which is taken as the origin. A \emph{rare configuration} consists of vectors $g_1,\dots,g_q\in\R^d$, the gradients of the rare rows, with coefficients, or \emph{rates}, $a_i>0$, their counts in a proportional selection. The \emph{target} $r$, which is the fit of the full dataset in these coordinates, the \emph{energy} $C$ of the rare rows, the matrix $S$ and the total $A$ are
\[
 r=\sum_ia_ig_i,\qquad C=\sum_ia_i\|g_i\|^2,\qquad S=\sum_ia_ig_ig_i^\top,\qquad A=\sum_ia_i .
\]
The selector chooses counts $k\in\Z_{\ge0}^q$ and a density $1+h\ge0$ on the types with $\E_\nu h=0$ such that $G=\E_\nu[(1+h)xx^\top]$ is positive definite. The \emph{fit} is the solution $w$ of $Gw=\sum_ik_ig_i+\E_\nu[(1+h)\phi x]$, and the \emph{cost} is $\|w-r\|^2$; the infimum of the cost is the \emph{value} of the game. For $h=0$ the cost is $\|\delta\|^2$ with $\delta=\sum_i(k_i-a_i)g_i$, so that the infimum of the cost is at most $\Lambda_dC$: this is plain rounding. Write
\[
 s_x=\langle x,r\rangle-\phi_x,\qquad n_x=\langle x,\delta\rangle,\qquad \ell_x=s_x+n_x .
\]
Up to sign, $s_x$ is the residual of the type $x$ at the target, $\ell_x$ is its residual at the plain fit $r+\delta$, and $n_x$ is the effect of the rounding noise on it. The index $x$ is often dropped inside averages, so that $n\ell$ there means $n_x\ell_x$; elsewhere $n$ is the budget. Without indices, $A$ is the total rate $\sum_ia_i$; the minimax values carry indices, except in the shorthand $B/A$ of Section~\ref{sec:intro}. A direct computation gives the \emph{exact formula}
\begin{equation}\label{eq:fitmain}
 w-r=\delta-(I+H)^{-1}b,\qquad H=\E_\nu[hxx^\top],\quad b=\E_\nu[h\,\ell_x\,x].
\end{equation}
Since the types are isotropic, $\|\E_\nu[ux]\|^2\le\E_\nu u^2$ for every function $u$. This estimate, which does not depend on the dimension, is the reason why dimension-free bounds are possible.

\begin{theorem}\label{thm:U}
Let $J=10^{2364}$. For every $d$, every affine frame in $\R^d$ and every rare configuration there are $k$ and $h$ with $\sum_ik_i\le\lceil J(1+A)\rceil$, $-\frac12\le h\le4$ and cost at most $(1-J^{-1})\,C$.
\end{theorem}

We put $c_*=1-J^{-1}$. The point of the theorem is that $c_*<1$ does not depend on the dimension. Section~\ref{sec:one} outlines the proof, and Appendix~\ref{app:one} contains it.

\section{One label}\label{sec:one}

\begin{theorem}[One label]\label{thm:one}
With the absolute constant $c_*=1-10^{-2364}$ of Section~\ref{sec:game}, for all $d\ge1$ and $\kappa\ge1$,
\[
 \tfrac12\ \le\ \underline\alpha^{(1)}_{d,\kappa}\ \le\ \overline\alpha^{(1)}_{d,\kappa}\ \le\ \min\{\Lambda_d,c_*\}.
\]
In particular $\lim_nn\,A^{(1)}_{d,\kappa}(n)=d\kappa/2$ for $d\le4$.
\end{theorem}

The lower bound uses one rare row of maximal leverage along $\be_1$, a direction in which all frequent rows have the same coordinate, with mass $1/(2n)$, and the bound $\Lambda_d$ is plain rounding (both in Appendix~\ref{app:aware}). The bound $c_*$ follows from Theorem~\ref{thm:U} through Theorem~\ref{thm:data} in Appendix~\ref{app:transfer}: for the frequent rows of a dataset, $h\ge-\frac12$ keeps the selected design well conditioned and the bounded counts make all errors uniform, so that the counts and the density given by Theorem~\ref{thm:U} are realized by a selection up to an error that vanishes when first $n\to\infty$ and then the threshold $\eta$ for rare rows tends to zero.

\paragraph{Outline of the proof of Theorem~\ref{thm:U}.} Let $C=1$, $\varepsilon=J^{-1}$, and suppose that each of the strategies (S1) to (S4) of Appendix~\ref{app:strategies}, whose counts are conditioned on $\sum_ik_i\le\lceil J(1+A)\rceil$, has mean cost above ${1-\varepsilon}$. Below, $k$ denotes one unconditioned draw with independent coordinates $k_i\sim\Pois(a_i)$, so that $\E\|\delta\|^2=1$. (S1) keeps the better of two conditioned draws with $h=0$. Its failure gives ${\E\big|\|\delta\|^2-1\big|}=O(\varepsilon)$. Removing the conditioning leaves the mean costs of (S3) and (S4) at least $1-2\varepsilon$. Below, $\approx$ and ``small'' refer to the explicit parameters \eqref{eq:params}.

\emph{Step 1: no first-order gain (Lemmas~\ref{lem:gain}, \ref{lem:shell}).} Let $\omega_x=\min\{1,|n_x|/|\ell_x|\}$, and $\omega_x=1$ if $\ell_x=0$. If $|\psi|\le\omega$ and $\E_\nu\psi=0$, then \eqref{eq:fitmain} with $h=\xi\psi$ and small $\xi>0$ gives cost at most $\|\delta\|^2-2\xi\,\E_\nu[\psi\,n\ell]+O(\xi^2)\|\delta\|^2$: the weight $\omega$ keeps the quadratic term bounded however large the residuals are. The best $\psi$ has $\E_\nu[\psi\,n\ell]=\min_c\E_\nu[\omega\,|n\ell-c|]$. (S2) uses it and fails. Hence, for most $k$, $n_x\ell_x$ is nearly one number $c=c(k)$, in a weighted average over the types.

\emph{Step 2: high rank (Lemma~\ref{lem:rank}).} Split $\delta=X+Y$ into two independent Poisson halves. Averaging $\|\delta\|^2\approx1$ over one half gives $\|X\|^2\approx\|Y\|^2\approx\frac12$, hence $\langle X,Y\rangle\approx0$ and, by independence and a truncation, $\tr S^2=4\,\E\langle X,Y\rangle^2\approx0$: the noise is spread over many directions. Conditioning on the event that the longer rare gradients are not selected shows that these carry little energy.

\emph{Step 3: rare noise (Lemmas~\ref{lem:conc} to~\ref{lem:arrival}).} For each type, $n_x$ is a centered compound Poisson variable with variance $\sigma_x^2=x^\top Sx$. Draw two types $x,y$ independently with probabilities $\nu\sigma^2$. By Step~2, bounded Lipschitz tests of $(n_x/\sigma_x,n_y/\sigma_y)$ are close, on average over the two types, to the same tests for independent coordinates. Step~1 controls weighted discrepancies of $n_x\ell_x$ from the common number $c$. Comparing two types eliminates $c$ (see \eqref{eq:pair}): the difference $n_x\ell_x-n_y\ell_y$, divided by a common scale and restricted to values of $n_x/\sigma_x$ and $n_y/\sigma_y$ that are bounded away from zero and infinity, is small on average over the pairs. But $n_x$ is the sum of three independent pieces with the same law. If $n_x/\sigma_x$ is not small, replacing the pieces one at a time by independent copies produces, with a probability that is bounded below, four well-separated values of $n_x/\sigma_x$. At least three of them are not close to zero, and the quadratic polynomial $n_x(s_x+n_x)$, divided by its scale $\sigma_x^2\max\{1,|s_x|/\sigma_x\}$, cannot be close to one number at all three. Hence $n_x/\sigma_x$ is small with high probability for most types. Its variance is one, so it is carried by rare jumps. Fix a small $\varrho>0$ and call the rare row $i$ \emph{large} at $x$ if $|\langle x,g_i\rangle|>\sigma_x/\sqrt\varrho$. The large rows have total rate $p_x\le\varrho$, and in integrated mean square $n_x$ is approximated by the effect of a unique large arrival.

\emph{Step 4: cancellation (Lemma~\ref{lem:cancel}).} (S3) changes the density of the types where a unique large arrival, of projection $g$, has $|n_x-g|\le|g|/2$, with the weight $\omega_x$ and in the direction of the gain, and compensates the mean by a multiple of a profile $m_x\le p_x$, whose own quadratic cost, averaged over the types and the counts, is at most $10\varrho$, whatever the size of $s_x$. Outcomes on which a bound for this multiple exceeds a cutoff are given up. The compensation is then nearly free, so $c$ may be taken to be zero, and the failure of (S3) forces $|\ell_x|\ll|n_x|$ after a large arrival: \emph{the jump is nearly equal to $-s_x$}, in an integrated squared-error sense; in this sense one selected rare row cancels the residual of the type.

\emph{Step 5: doubling (Lemma~\ref{lem:double}).} Where $\ell_x\approx0$ a change of density has no effect, so (S4) overshoots. It uses two independent streams of arrivals for each rare row, single copies at rate $a_i/2$ and double copies at rate $a_i/4$. The mean of the counts is unchanged and the plain cost rises to $\frac12+1=\frac32$, the doubles contributing $1$. On the types hit by a double of a row that is large there and whose single jump cancels there up to a small relative error $\gamma$, $n_x\approx-2s_x$ and $\ell_x\approx-s_x\approx n_x/2$ in integrated mean square, and their density is multiplied by $5$ unless the outcome is given up. With $P=\E_\nu[xx^\top;\text{double}]$, $0\preceq P\preceq I$, \eqref{eq:fitmain} gives the error $\approx[I-2(I+4P)^{-1}P]\delta$. By convexity its squared norm is at most $\|\delta\|^2-\frac{16}{25}\delta^\top P\delta$, and $\E\,\delta^\top P\delta\approx1$ is the energy of the doubles, which leaves $\frac32-\frac{16}{25}<\frac78$. Giving up the outcomes with too large a set of such types, of probability at most $\frac1{64}$, leaves a mean cost of at most about $\frac78$. With the chosen parameters it is below $0.99<1-2\varepsilon$, a contradiction.

\emph{Constants.} Steps 4 and 5 fix $\gamma=\frac9{2000}$, the accuracy of the cancellation, then the cutoff and the step size, then $\varrho=2\cdot10^{-33}$ and the level $\theta=5\cdot10^{-24}$ of ``small'' in Step~3. Eliminating $c$ in Step~3 uses $11\varepsilon\le10^{-126}\theta^{96}$, which gives $J=10^{2364}$.

\medskip
The strategies also yield the following algorithm.

\begin{theorem}[Efficient selection]\label{thm:algo}
There are an absolute constant $c<1$ and a randomized algorithm with the following property. Given a dataset $D$ with $N$ rows, $d$ features, one label per row, $L^*_D>0$ and $\kappa_D\le\kappa$, a budget $n$ and $\epsilon\in(0,1)$, it uses a number of arithmetic operations, comparisons and random bits that is polynomial in $N$, $d$, $n$ and $\log(1/\epsilon)$, in the real-arithmetic model with an oracle that returns optimal basic solutions of linear programs, and returns a selection with
\[
 n\,\rho\le\big(c+O_{d,\kappa}(n^{-1/16})\big)\,d\kappa\qquad\text{with probability at least }1-\epsilon,
\]
for all $n\ge n_0(d,\kappa)$, uniformly over the datasets.
\end{theorem}

The algorithm samples the counts of the rare rows, sometimes doubled, perturbs the proportions of the frequent rows by the rules of Steps 1, 4 and 5, and keeps the best selection found. It does not need to know $c$. The theorem is a worst-case asymptotic guarantee. The proof gives ${1-c}=\frac12\,10^{-2364}$ and requires $\log_2(n+1)$ to be at least of order $10^{2364}$, and we do not claim that the algorithm is preferable to simple rounding of the counts on a given dataset. The proof is in Appendix~\ref{app:algo}.

\section{A barrier for one label in high dimension}\label{sec:barrier}

\begin{theorem}\label{thm:barrier}
Let $m=1$.
\begin{enumerate}
\item[(i)] For every $c<e^{-1/2}$ there is a $d_1$ such that $\underline\alpha^{(1)}_{d,\kappa}>c$ for all $d\ge d_1$ and all $\kappa>1$.
\item[(ii)] $\underline\alpha^{(1)}_{d,\kappa}>0.5006$ whenever $\kappa>1$, $d\ge332\,101$ and $d\kappa\ge332\,100\,000$.
\end{enumerate}
\end{theorem}

\paragraph{Outline of the proof.} The frequent rows are multiples $\be_j/\sqrt{\nu_j}$ of the unit vectors, so in the limiting game reshaping multiplies each coordinate of the selected gradient by a positive factor. One block of them, the \emph{heavy} types, has mass close to one; hence the factors of the heavy coordinates cannot all be much smaller than one. All other types have small mass. There are $M$ rare rows with the same small coefficient $a$, of total $A=Ma$, and their gradients have three kinds of coordinates. A coordinate is \emph{lost} if the selected gradient vanishes there or has the wrong sign; its squared target is then part of the cost, whatever the proportions. In the heavy coordinates all rare gradients agree and have positive entries. This \emph{caps} the total count: a count that exceeds $A$ by a fixed factor is too expensive. Each rare row has a \emph{poison} coordinate, in which its own entry is strongly negative and all others are slightly positive; it is lost if the row is selected and the total count is below the cap. Each set in a family of sets of rare rows has a \emph{cover} coordinate, in which the rows of the set have equal positive entries and the others vanish; it is lost unless a row of the set is selected. Let the poison coordinates have squared target $\beta$ each, and let the covers be all sets of about $\sqrt e/a$ rows, with total squared target $\sqrt e\,\beta A$. Poisons and covers then have energy about $\beta A$ each, and the heavy energy is negligible. If the total count is below the cap and $xA$ distinct rows are selected, the cost is at least about $\beta A\,(x+\sqrt e\,e^{-\sqrt e\,x})$. The minimum, at $x=e^{-1/2}$, is $2e^{-1/2}\beta A$, the fraction $e^{-1/2}$ of the energy. In both parts the dimension is filled by enlarging the heavy block; for part~(ii) the covers are built from a Steiner system $S(3,10,82)$ and the numerical inequalities are verified in exact arithmetic. The proof is in Appendix~\ref{app:barrier}.

\medskip
The smallest dimension in which $\underline\alpha^{(1)}_{d,\kappa}>\frac12$ for some $\kappa$ is not known; it is at least $5$ and at most $332\,101$. If all total rare counts up to $A$ may be used, some selection loses poison and cover coordinates whose squared targets sum to at most the fraction $e^{-1/2}+O(1/A)$ of the energy of all these coordinates (Proposition~\ref{prop:limits} in Appendix~\ref{app:limits}); this does not bound the value of the game.

\section{Several labels}\label{sec:several}

We return to $m\ge1$ labels per row.

\begin{theorem}[Several labels]\label{thm:several}
For all $d\ge1$, $m\ge2$ and $\kappa\ge1$,
\[
 \Gamma_{d,m-1}\ \le\ \underline\alpha^{(m)}_{d,\kappa}\ \le\ \overline\alpha^{(m)}_{d,\kappa}\ \le\ \Gamma_{d,m}.
\]
If $dm\le4$, then $\lim_nn\,A^{(m)}_{d,\kappa}(n)=d\kappa/2$, and if $d=1$, then $\lim_nn\,A^{(m)}_{1,\kappa}(n)=\kappa\Lambda_m$. For fixed $m\ge2$, by the asymptotic formula for $\Lambda_D$ in Section~\ref{sec:setting}, both constants are $1-\Theta(d^{-1/3})$ as $d\to\infty$.
\end{theorem}

The upper bound is plain rounding, with the rank-one structure of the gradients retained, and holds for $m=1$ as well. The statements for $dm\le4$ and for $d\to\infty$ follow from $\frac12\le\Lambda_d\le\Gamma_{d,m-1}\le\Gamma_{d,m}\le\Lambda_{dm}$, from $\Lambda_{dm}=\frac12$ for $dm\le4$ (Proposition~\ref{prop:Lambda}), and from the asymptotic formula for $\Lambda_D$. By Proposition~\ref{prop:Lambda}(iv) the lower leading constant exceeds $\frac12$ whenever $d(m-1)\ge5$, and the upper bound $\Gamma_{d,m}$ equals $\frac12$ only if $dm\le4$ (compare Theorem~\ref{thm:barrier}). The upper bound and the case $d=1$ are in Appendix~\ref{app:aware}. We prove the lower bound here.

The following lemma perturbs the simplex so that the second moment of the whole dataset is exactly $I$.

\begin{lemma}\label{lem:frame}
Let $z_0,\dots,z_d\in\R^d$ be the vertices of a regular simplex with $\|z_j\|^2=d$, so that $\sum_jz_j=0$ and $\frac1{d+1}\sum_jz_jz_j^\top=I$. For every symmetric $F$ close enough to $I$ there are vectors $x_j=x_j(F)$, continuous in $F$, with $x_j(I)=z_j$, $\frac1{d+1}\sum_jx_jx_j^\top=F$ and $\|x_j\|^2=\tr F$ for all $j$. They satisfy a unique linear relation $\sum_jh_jx_j=0$ with $\sum_jh_j=d+1$, and $h_j\to1$ as $F\to I$.
\end{lemma}

\begin{proof}
For $d=1$ take $\pm\sqrt F$. Let $d\ge2$ and first $\tr F=d$. Consider the map $(x_j)\mapsto\frac1{d+1}\sum_jx_jx_j^\top$ from the product of $d+1$ spheres of radius $\sqrt d$ to the symmetric matrices of trace $d$. If a traceless symmetric $T$ is orthogonal to the range of its derivative at $(z_j)$, then $Tz_j\parallel z_j$ for every $j$. Eigenvectors of a symmetric matrix with different eigenvalues are orthogonal, and $\langle z_i,z_j\rangle=-1\ne0$, so all eigenvalues agree, $T$ is a multiple of $I$, and $T=0$. The derivative is onto and the inverse function theorem applies on a complement of its kernel. In general apply this to $dF/{\tr F}$ and scale. Near the simplex $(x_j)$ has rank $d$, so its linear relations form a line that depends continuously on $F$ and is spanned by $(1,\dots,1)$ at $F=I$; this gives the last statement.
\end{proof}

\begin{proof}[Proof of the lower bound in Theorem~\ref{thm:several}]
Put $p=m-1$ and $R=d\kappa$. By Proposition~\ref{prop:Lambda} it suffices to treat linearly independent matrices $g_i=\xi_iu_i^\top\in\R^{d\times p}$, $i\le q$, normalized so that $\|\xi_i\|^2=R$, and rational $a_i\in(0,1)$. Let $s=\sum a_i$, $M=\sum a_ig_i$, $C=\sum a_i\|g_i\|_F^2=R\sum a_i\|u_i\|^2$, and fix $\varepsilon>0$. For large $n$ let
\[
 F=\frac{I-\frac1n\sum_ia_i\xi_i\xi_i^\top}{1-s/n},\qquad b=\frac{F^{-1}M}{1-s/n}\in\R^{d\times p},
\]
and $x_j=x_j(F)$, $h_j$ as in Lemma~\ref{lem:frame}. The dataset has, for each $i$, a \emph{rare} row with feature $\xi_i$, first $p$ labels $\sqrt n\,u_i$ and last label $0$, of mass $a_i/n$; and, for each $0\le j\le d$, a \emph{frequent} row with feature $x_j$, first $p$ labels $-b^\top x_j/\sqrt n$ and last label $\varepsilon h_j$, of mass $(1-s/n)/(d+1)$. All masses are rational. The second moment is exactly $I$; the rare rows have squared norm $d\kappa$ and the frequent rows $\tr F=d(1-\kappa s/n)/(1-s/n)\le d$, so the normalized leverage is $\kappa$ for every $\kappa\ge1$. The gradient of the full loss vanishes at zero: for the first $p$ labels this is $M/\sqrt n-(1-s/n)Fb/\sqrt n=0$, and for the last it is $\sum_jh_jx_j=0$. Hence $W^*=0$ and $L^*\to C/R+\varepsilon^2$.

If $\liminf_nn\min_k\rho(k)=\infty$, there is nothing to prove. Otherwise consider, along a subsequence that attains the lower limit, optimal selections with rare counts $k_i$ and frequent counts $c_j$; for them $n\rho$ is bounded. Since $\Sigma=I$, the fit is $O(n^{-1/2})$, whether or not the design is singular, and it satisfies the normal equations with a design $S_n$ with $\|S_n\|\le nR$. The first $p$ normal equations, divided by $\sqrt n$, show that $\sum_ik_ig_i=O(1)$, because the first labels of the frequent rows are $O(n^{-1/2})$. By linear independence the $k_i$ are bounded, and after passing to a further subsequence they are constant and the $c_j/n$ converge. The last normal equation, divided by $n$, reads $\varepsilon\sum_j(c_j/n)h_jx_j=O(n^{-1/2})$, while $\sum_jc_j/n\to1$. In the limit the only relation among the $z_j$ is the constant one, so $c_j/n\to1/(d+1)$: \emph{the last label freezes the design}. Hence $S_n/n\to I$, and the first $p$ normal equations give $\sqrt n\,\hat W_{1:p}^\top\to\sum_i(k_i-a_i)g_i$. Therefore $\liminf_nn\min\rho\ge\Lambda(g,a)\,C/(C/R+\varepsilon^2)$ for these datasets. Let $\varepsilon\to0$ and take the supremum over configurations.
\end{proof}

\paragraph{The value of labels.} We compare the two information models.

\begin{corollary}[Value of labels]\label{cor:value}
For all large $n$ let $V^{(m)}_{d,\kappa}(n)=B^{(m)}_{d,\kappa}(n)/A^{(m)}_{d,\kappa}(n)$, whose numerator and denominator are then finite and positive.
\begin{enumerate}
\item[(i)] If $m=1$, then $1/{\min\{\Lambda_d,c_*\}}\le\liminf_nV^{(1)}_{d,\kappa}(n)\le\limsup_nV^{(1)}_{d,\kappa}(n)\le2$ for all $d$ and $\kappa$. If $\kappa>1$, then $\limsup_{d}\limsup_nV^{(1)}_{d,\kappa}(n)\le\sqrt e$.
\item[(ii)] If $m\ge2$, then $1/\Gamma_{d,m}\le\liminf_nV^{(m)}_{d,\kappa}(n)\le\limsup_nV^{(m)}_{d,\kappa}(n)\le1/\Gamma_{d,m-1}$ for all $d$ and $\kappa$, and both bounds are $1+\Theta(d^{-1/3})$ as $d\to\infty$ with $m$ fixed.
\item[(iii)] If $dm\le4$, then $\lim_nV^{(m)}_{d,\kappa}(n)=2$.
\end{enumerate}
\end{corollary}

The corollary follows from Theorems~\ref{thm:blind}, \ref{thm:one}, \ref{thm:barrier} and~\ref{thm:several}.

\section{Grouped designs}\label{sec:group}

In this section $m=1$. A dataset or design is \emph{grouped} if every feature vector is one of the unit vectors $\be_1,\dots,\be_d$; group $j$ consists of the rows with feature $\be_j$. The fit is then the vector of the means of the selected labels in the groups, with zero for a group without selected rows, and reshaping is the allocation of a common budget. Let $A^{\mathrm{grp}}_d(n)$ and $B^{\mathrm{grp}}_d(n)$ be defined as in Definition~\ref{def:AB}, over grouped datasets and designs with $d$ nonempty groups and no restriction on the leverage.

\begin{theorem}[Grouped designs]\label{thm:group}
For every $d\ge1$: (i) $n\,A^{\mathrm{grp}}_d(n)>27d/128$ for every $n\ge1$; (ii) $\limsup_nn\,A^{\mathrm{grp}}_d(n)\le27d/128+13/24$ for $d\ge2$, and $\lim_nn\,A^{\mathrm{grp}}_1(n)=\frac12$; (iii) ${\lim_nn\,B^{\mathrm{grp}}_d(n)=d}$.
\end{theorem}

\paragraph{Proof of (i) and outline of the proof of (ii).} For $n<d$ every selection leaves some group empty; if every group has mass $1/d$ and the labels $\mu\pm1$ in equal numbers, the excess ratio is then at least $\mu^2/d$, so the value is infinite. For $n\ge d$ give all groups equal mass and let a fraction $p=27d/(64n)$ of the labels in each group equal one, the others being zero, so that $L^*=p(1-p)$. A group that receives $b\ge1$ rows, of which $k$ carry the rare label, has squared error $p^2(k/t-1)^2$ with $t=bp$, an empty group has squared error $p^2$, and
\begin{equation*}
 \Big(\frac kt-1\Big)^2\ \ge\ 1-\frac{32}{27}\,t\qquad\text{for all }t>0\text{ and all integers }k\ge0 .
\end{equation*}
This is clear if $k=0$ or $t\ge27/32$; otherwise $k/t-1\ge1/t-1>0$ and $(1/t-1)^2-1+\frac{32}{27}t=(4t-3)^2(2t+3)/(27t^2)\ge0$. Summing over the groups, every selection has excess loss at least $p^2(1-\frac{32np}{27d})=\frac{p^2}2$, so that $n\,\rho\ge np/(2(1-p))>27d/128$. For (ii) with $d\ge2$ each group is reduced to one or two labels. Groups whose rarer label has weight below $n^{-3/4}$ are given a target number of copies of it equal to $0$, $\frac34$ or $1$, up to rounding, and initially one copy in the last two cases, the count being rounded again when spare rows are added; the other groups need only $o(n)$ rows. A Lagrangian bound for the allocation problem is at most $27d/128$ at unit budget, and the greedy solution loses at most the effect of one step. The proofs of (ii) and (iii) are in Appendix~\ref{app:group}.

\section{Discussion}\label{sec:disc}

We determined the leading constant of unweighted data selection for least squares under a leverage bound for label-blind selectors in all dimensions and for label-aware selectors when $dm\le4$ or $d=1$. With one label per row the gain from the labels is bounded away from a factor of one, uniformly in $d$ and $\kappa$; for any fixed $m\ge2$ it tends to a factor of one as $d$ grows. The following questions remain open.

\emph{One label.} The main open problem is the value of $\lim_nn\,A^{(1)}_{d,\kappa}(n)/(d\kappa)$, or even its existence, for $d\ge5$. Let $c_{\mathrm{sc}}=\sup_{d,\kappa}\overline\alpha^{(1)}_{d,\kappa}$, so that $e^{-1/2}\le c_{\mathrm{sc}}\le c_*=1-10^{-2364}$. Is $c_{\mathrm{sc}}=e^{-1/2}$? The explicit value of $c_*$ is extremely weak, and a proof of $c_{\mathrm{sc}}\le0.99$, say, would be interesting. In particular we do not know whether the lower leading constant equals $\frac12$ for $d=5$. Theorem~\ref{thm:barrier} requires $\kappa>1$, and we do not know whether $\underline\alpha^{(1)}_{d,1}>\frac12$ for some $d$.

\emph{Several labels.} For $d,m\ge2$ with $dm\ge5$, does the limit exist, and is $\underline\alpha^{(m)}_{d,\kappa}=\Gamma_{d,m-1}$, is $\overline\alpha^{(m)}_{d,\kappa}=\Gamma_{d,m}$, or neither? The constants $\Gamma_{d,p}$ with $d,p\ge2$ and $dp\ge5$ exceed $\frac12$ but are not known.

\appendix\clearpage

\section{Notation}\label{app:notation}

Table~\ref{tab:notation} lists the main symbols that are used in several sections. Other symbols are explained where they are used.

\begin{table}[ht]
\centering\small
\renewcommand{\arraystretch}{1.15}
\begin{tabular}{@{}l@{\quad}p{0.63\linewidth}@{\ \ }l@{}}
\toprule
Symbol & Meaning & Defined in\\
\midrule
$d$, $m$, $N$ & features, labels per row, rows of the dataset $D$ & Section~\ref{sec:setting}\\
$n$ & budget: number of selected rows, with repetition & Section~\ref{sec:setting}\\
$\kappa_D$, $\kappa$ & normalized leverage $\max_ix_i^\top\Sigma_D^{-1}x_i/d$; upper bound for it & Section~\ref{sec:setting}\\
$k$ & selection: $k\in\Z_{\ge0}^N$ with $\sum_ik_i=n$ & Section~\ref{sec:setting}\\
$\rho_D(k)$ & excess ratio $(L_D(\hat W_k)-L_D^*)/L_D^*$ & Section~\ref{sec:setting}\\
$A^{(m)}_{d,\kappa}(n)$ & label-aware minimax ratio & Definition~\ref{def:AB}\\
$B^{(m)}_{d,\kappa}(n)$ & label-blind minimax expected ratio & Definition~\ref{def:AB}\\
$\underline\alpha^{(m)}_{d,\kappa}$, $\overline\alpha^{(m)}_{d,\kappa}$ & $\liminf_n$, $\limsup_n$ of $n\,A^{(m)}_{d,\kappa}(n)/(d\kappa)$ & Section~\ref{sec:intro}\\
$\Lambda_D$ & rounding constant of vectors in $\R^D$ (here $D$ is a dimension) & Section~\ref{sec:setting}\\
$\Gamma_{d,p}$ & rounding constant of $d\times p$ matrices of rank at most one & Section~\ref{sec:setting}\\
$x$, $\nu$, $\phi$ & frame: types $x\in\R^d$ with masses $\nu$ and scaled residuals $\phi_x$ & Section~\ref{sec:game}\\
$g_i$, $a_i$ & gradients of the rare rows and their coefficients $a_i>0$ & Section~\ref{sec:game}\\
$r$, $C$ & target $\sum_ia_ig_i$; rare energy $\sum_ia_i\|g_i\|^2$ & Section~\ref{sec:game}\\
$A$, $S$ & $\sum_ia_i$ (no indices); $\sum_ia_ig_ig_i^\top$ & Section~\ref{sec:game}\\
$k_i$, $h$ & counts of the rare rows; density $1+h\ge0$ on the types, $\E_\nu h=0$ & Section~\ref{sec:game}\\
$G_k$; $G$ & design $\sum_ik_ix_ix_i^\top$; in the game $\E_\nu[(1+h)xx^\top]$ & Sections~\ref{sec:setting}, \ref{sec:game}\\
$w$ & fit: $Gw=\sum_ik_ig_i+\E_\nu[(1+h)\phi x]$, with cost $\|w-r\|^2$ & Section~\ref{sec:game}\\
$\delta$ & rounding noise $\sum_i(k_i-a_i)g_i$ & Section~\ref{sec:game}\\
$s_x$ & $\langle x,r\rangle-\phi_x$ & Section~\ref{sec:game}\\
$n_x$, $\ell_x$ & $\langle x,\delta\rangle$ (not the budget); $s_x+n_x$ & Section~\ref{sec:game}\\
$V^{(m)}_{d,\kappa}(n)$ & value of labels $B^{(m)}_{d,\kappa}(n)/A^{(m)}_{d,\kappa}(n)$ & Section~\ref{sec:several}\\
$J$, $c_*$ & $J=10^{2364}$ (Theorem~\ref{thm:U}); $c_*=1-J^{-1}<1$ & Section~\ref{sec:game}\\
\bottomrule
\end{tabular}
\caption{Main notation. In Appendix~\ref{app:barrier}, $m$ denotes a total count of rare rows, $N$ a threshold for it and $h$ a number of coordinates.}\label{tab:notation}
\end{table}

\section{Proof of Proposition~\ref{prop:Lambda}}\label{app:round}

\begin{lemma}\label{lem:conic}
If $\Lambda(g,a)>0$, there are a linearly independent subfamily $g_T=(g_i)_{i\in T}$ and $f\in[0,1)^T$ with $C(g_T,f)>0$ and $\Lambda(g,a)\le\Lambda(g_T,f)$.
\end{lemma}

\begin{proof}
We have $r=\sum_ia_ig_i\ne0$, since otherwise $k=0$ is exact. Delete the zero vectors. The linear program of minimizing $\sum_ib_i\|g_i\|^2$ over $b\ge0$ with $\sum_ib_ig_i=r$ is feasible and has positive costs, so it has an optimal basic solution $b$; its support $T$ indexes linearly independent vectors, and $\sum_Tb_i\|g_i\|^2\le C(g,a)$. Put $f=b-\lfloor b\rfloor$. The counts $k'+\lfloor b\rfloor$ on $T$, with $k'\in\Z_{\ge0}^T$, are admissible for $(g,a)$ and have the error $\sum_T(k'_i-f_i)g_i$. If $C(g_T,f)=0$, then $k'=0$ is exact and $\Lambda(g,a)=0$, a contradiction. Otherwise $\Lambda(g,a)\,C(g,a)\le\Lambda(g_T,f)\,C(g_T,f)\le\Lambda(g_T,f)\,C(g,a)$.
\end{proof}

\begin{proof}[Proof of Proposition~\ref{prop:Lambda}]
(i) One unit vector with $a=\frac12$ gives $\Lambda=\frac14/\frac12$.
Embedding $\R^D$ in $\R^{D+1}$ gives monotonicity. By Lemma~\ref{lem:conic} we may assume that $a\in[0,1)^q$ and that the $g_i$ are linearly independent; zero coefficients may be deleted, and for linearly independent vectors $\Lambda(g,a)C(g,a)$ is the squared distance from $\sum a_ig_i$ to a closed discrete set and hence continuous in $a$, so that rational coefficients suffice. The lemma keeps a subfamily of the given vectors, and therefore applies to rank-one matrices. Independent Bernoulli counts $k_i$ with means $a_i$ have the mean squared error $\sum_ia_i(1-a_i)\|g_i\|^2\le C(g,a)$, and some outcome is at most the mean; hence $\Lambda_D\le1$.

(ii) By Lemma~\ref{lem:conic} we may assume that there are at most four vectors and that $a_i<1$. Let $I$ be the set of indices with $a_i<\frac12$, $s=\sum_Ia_i<2$ and $v=\sum_Ia_ig_i$. For indices outside $I$ take independent Bernoulli counts with means $a_i$. On $I$, independently of these, take $k=\be_i$ with probability $a_i/2$ for each $i\in I$, and $k=0$ with probability $1-s/2$. The mean squared error of the second part is
\[
 \sum_I\frac{a_i}2\|g_i-v\|^2+\Big(1-\frac s2\Big)\|v\|^2=\frac12\sum_Ia_i\|g_i\|^2 ,
\]
and the first part has mean error zero and mean squared error $\sum_{i\notin I}a_i(1-a_i)\|g_i\|^2$, which is at most $\frac12\sum_{i\notin I}a_i\|g_i\|^2$. Hence the total mean squared error is at most $C(g,a)/2$, and with (i), $\Lambda_D=\frac12$.

(iii) Matrices in $\R^{d\times1}$ have rank at most one, transposition preserves rank and norm, vectors $g_i\in\R^d$ give the rank-one matrices $g_iu^\top$ with a fixed unit vector $u\in\R^p$, and $\R^{d\times p}$ is a Euclidean space of dimension $dp$. Appending a zero row or column preserves rank, norms and rounding costs, which gives monotonicity in each index.

(iv) If $dp\le4$, then $\frac12\le\Gamma_{d,p}\le\Lambda_{dp}=\frac12$ by (i) to (iii). For the converse let $Q=(6I+\mathbf J)/7$, where $\mathbf J$ is the $5\times5$ matrix of ones, let $g_1,\dots,g_5$ be elements of a Euclidean space with Gram matrix $Q$, and let $a_i=\frac9{22}$ for all $i$. For $k\in\Z_{\ge0}^5$ and $K=\sum_ik_i$,
\begin{align*}
 \Big\|\sum_i(k_i-a_i)g_i\Big\|^2&=\frac67\sum_i(k_i-a_i)^2+\frac17\Big(K-\frac{45}{22}\Big)^2\\
 &=\frac{317}{308}+\frac67\sum_ik_i(k_i-1)+\frac17(K-1)(K-2)\ \ge\ \frac{317}{308},
\end{align*}
and $C(g,a)=\frac{45}{22}$, so that $\Lambda(g,a)\ge\frac{317}{630}>\frac12$. Since $Q\succ0$, such $g_i$ exist in $\R^5$; hence $\Gamma_{1,p}=\Gamma_{p,1}=\Lambda_p>\frac12$ for $p\ge5$. We show that they also exist among the matrices of rank one in $\R^{2\times3}$; by (iii) this covers all other cases with $dp\ge5$. Let $u,v=(6\pm\sqrt{22})/14\in(0,1)$, so that $u+v=\frac67$, $uv=\frac1{14}$ and $(1-u)(1-v)=\frac3{14}$, and let
\[
 x_1=(1,0),\qquad x_{2,3}=(\sqrt u,\pm\sqrt{1-u}),\qquad x_{4,5}=(\sqrt v,\pm\sqrt{1-v}) .
\]
These are unit vectors in $\R^2$ with $\langle x_2,x_3\rangle=2u-1=(\sqrt{22}-1)/7$, $\langle x_4,x_5\rangle=2v-1=-(\sqrt{22}+1)/7$, and $\langle x_2,x_4\rangle=\langle x_3,x_5\rangle$ and $\langle x_2,x_5\rangle=\langle x_3,x_4\rangle$ have the sum $2\sqrt{uv}$ and the product ${uv-(1-u)(1-v)}=-\frac17$; in particular no inner product vanishes. Let $B_{ii}=1$ and $B_{ij}=1/(7\langle x_i,x_j\rangle)$ for $i\ne j$. In the orthonormal basis $\be_1$, $(\be_2+\be_3)/\sqrt2$, $(\be_4+\be_5)/\sqrt2$, $(\be_2-\be_3)/\sqrt2$, $(\be_4-\be_5)/\sqrt2$ the matrix $B$ is the direct sum of
\[
 B_+=\begin{pmatrix}1&\alpha&\beta\\ \alpha&\frac{22+\sqrt{22}}{21}&-\sqrt{2/7}\\ \beta&-\sqrt{2/7}&\frac{22-\sqrt{22}}{21}\end{pmatrix}
 \qquad\text{and}\qquad
 B_-=\begin{pmatrix}\frac{20-\sqrt{22}}{21}&\sqrt{6/7}\\ \sqrt{6/7}&\frac{20+\sqrt{22}}{21}\end{pmatrix},
\]
where $\alpha=\sqrt2/(7\sqrt u)$ and $\beta=\sqrt2/(7\sqrt v)$, so that $\alpha^2=\frac2{49}(6-\sqrt{22})$, $\beta^2=\frac2{49}(6+\sqrt{22})$ and $\alpha\beta=\frac{2\sqrt{14}}{49}$. We have $\det B_-=\frac{400-22}{441}-\frac67=0$, and the Schur complement of the first entry of $B_+$ is
\[
 \begin{pmatrix}\frac{118+13\sqrt{22}}{147}&-\frac{9\sqrt{14}}{49}\\[2pt] -\frac{9\sqrt{14}}{49}&\frac{118-13\sqrt{22}}{147}\end{pmatrix},
 \qquad\text{with determinant}\quad\frac{118^2-169\cdot22}{147^2}-\frac{81\cdot14}{49^2}=0 .
\]
Both $2\times2$ matrices have positive diagonal entries and determinant zero, so they are positive semidefinite of rank one. Hence $B$ is positive semidefinite of rank three, and there are unit vectors $y_1,\dots,y_5\in\R^3$ with Gram matrix $B$. The matrices $g_i=x_iy_i^\top$ have rank one and $\langle g_i,g_j\rangle=\langle x_i,x_j\rangle\langle y_i,y_j\rangle=Q_{ij}$.
\end{proof}

\section{Proof of Theorem~\ref{thm:blind}}\label{app:blind}

\subsection{Upper bound}

Let $X$ be a design with normalized leverage at most $\kappa$. Whiten it;
the selections below have nonsingular designs, for which the excess ratio
is invariant. Let $i_1,\dots,i_n$ be independent and uniform on $[N]$, let
$k$ be the corresponding count vector, $G=\sum_tx_{i_t}x_{i_t}^\top$, and
for $\varepsilon\in(0,1)$ let $\mathcal E=\{G\succeq(1-\varepsilon)nI\}$. The
summands are independent, positive semidefinite, of norm at most $d\kappa$
and of mean $I$, so the matrix Chernoff bound
\citep[Theorem~1.1]{tropp2012}, with the elementary inequality
$e^{-\varepsilon}(1-\varepsilon)^{-(1-\varepsilon)}\le e^{-\varepsilon^2/2}$,
gives $\Prob(\mathcal E^c)\le d\exp(-\varepsilon^2n/(2d\kappa))$. Let
$n>2d\kappa\,\varepsilon^{-2}\log d$, so that $\Prob(\mathcal E)>0$, and let $Q$
be the law of $k$ conditional on $\mathcal E$; it depends on $X$, $n$ and
$\varepsilon$ only.

Let $Y$ be any label family with $L^*>0$ and let $e_i$ be the residuals of
the full fit. On $\mathcal E$, by \eqref{eq:defect}, the excess loss of $k$
is $\|G^{-1}b\|_F^2\le\|b\|_F^2/((1-\varepsilon)n)^2$ with
$b=\sum_tx_{i_t}e_{i_t}^\top$. The summands of $b$ are independent with mean
$N^{-1}\sum_ix_ie_i^\top=0$, so
$\E\|b\|_F^2=n\,N^{-1}\sum_i\|x_i\|^2\|e_i\|^2\le n\,d\kappa\,L^*$. Therefore
\[
 \E_{k\sim Q}\,\rho_{(X,Y)}(k)\le\frac{\E\|b\|_F^2}{(1-\varepsilon)^2n^2L^*\,\Prob(\mathcal E)}
 \le\frac{d\kappa}{n(1-\varepsilon)^2\big(1-d\,e^{-\varepsilon^2n/(2d\kappa)}\big)} ,
\]
uniformly in $X$ and $Y$. Letting $n\to\infty$ and then $\varepsilon\to0$
gives $\limsup_nn\,B^{(m)}_{d,\kappa}(n)\le d\kappa$; the choice
$\varepsilon^2=4d\kappa\log n/n$ gives the rate stated after Theorem~\ref{thm:blind}.

\subsection{Lower bound}

It suffices to take $m=1$: for a fixed design and a fixed law $Q$, scalar
label families padded with $m-1$ zero outputs are admissible, and padding
changes neither the excess loss of any selection nor $L^*$, so
$B^{(m)}_{d,\kappa}(n)\ge B^{(1)}_{d,\kappa}(n)$. Fix a rational
$p\in(0,1/(d\kappa))$ and put $c^2=(1-pd\kappa)/(1-p)\in(0,1]$ and
$\theta^2=(d-1)/(1-p)$. For an even integer $M$ such that $N=M/p$ is an
integer and, if $d\ge2$, $(1-p)N/(2(d-1))$ is an integer (there are
arbitrarily large such $M$), let $X$ consist of $M$ \emph{cluster rows}
equal to $\sqrt{d\kappa}\,\be_1$ and, for each $2\le j\le d$ and each sign,
of $(1-p)N/(2(d-1))$ rows $c\,\be_1\pm\theta \be_j$ (for $d=1$, of $(1-p)N$ rows
equal to $c$). Then $\Sigma_X=I$, the cluster rows have squared norm $d\kappa$ and the others $c^2+\theta^2=(d-pd\kappa)/(1-p)\le d\kappa$, so the normalized leverage of $X$ is $\kappa$. Every
non-cluster row has first coordinate $c$. For a selection $k$ let $s=s(k)$
be the number of selected cluster rows, counted with multiplicity, and
$\gamma(s)=sd\kappa+(n-s)c^2$, which is the entry $G_{11}$ of the selected
design $G$.

\emph{First label family.} For a vector $\sigma\in\{-1,1\}^M$ with zero sum
let $Y^\sigma$ assign $\sigma_l$ to the $l$th cluster row and zero to all
other rows. Then $\sum_ix_iy_i=0$, so $w^*=0$ and $L^*=p$. We claim that for
every selection $k$, if $\sigma$ is uniformly distributed,
\begin{equation}\label{eq:fam1}
 \E_\sigma\,\rho_{Y^\sigma}(k)\ \ge\ \frac{d\kappa\,s}{p\,\gamma(s)^2}\Big(1-\frac{n}{M-1}\Big).
\end{equation}
This is trivial for $s=0$. Let $s\ge1$ and let $k_l$ be the multiplicities
of the cluster rows. The selected gradient is $\sqrt{d\kappa}\,S\,\be_1$ with
$S=\sum_lk_l\sigma_l$, and the minimum-norm fit is
$\hat w=\sqrt{d\kappa}\,S\,G^+\be_1$. Since a cluster row is selected, $\be_1$
lies in the range of $G$, and for such vectors
$1=\langle (G^+)^{1/2}\be_1,G^{1/2}\be_1\rangle^2\le(\be_1^\top G^+\be_1)\,G_{11}$. Hence
$\|\hat w\|^2\ge\hat w_1^2\ge d\kappa S^2/\gamma(s)^2$. Finally
$\E\sigma_l\sigma_{l'}=-1/(M-1)$ for $l\ne l'$, so
$\E S^2=\sum_lk_l^2-(s^2-\sum_lk_l^2)/(M-1)\ge s-s^2/(M-1)$, which proves
\eqref{eq:fam1}.

\emph{Second label family.} Let $Y^1$ assign one to every cluster row and
zero to all other rows. Then
$L(w)=p(1-\sqrt{d\kappa}\,w_1)^2+\|w\|^2-pd\kappa w_1^2$, so
$w^*=p\sqrt{d\kappa}\,\be_1$ and $L^*=p(1-pd\kappa)>0$. For a selection $k$ the
selected gradient at zero is $b=s\sqrt{d\kappa}\,\be_1$, and every least-squares minimizer $\hat w$ satisfies $G\hat w=b$. Hence $G(\hat w-w^*)=b-Gw^*$, whose first coordinate equals
$\sqrt{d\kappa}\,(s-p\gamma(s))=\sqrt{d\kappa}\,c^2(s-np)$; here we used
$1-pd\kappa+pc^2=c^2$. On the other hand this coordinate is
$\langle G\be_1,\hat w-w^*\rangle$, and $\|G\be_1\|\le\sum_ik_i|x_{i1}|\,\|x_i\|\le n\,d\kappa$. Therefore
\begin{equation}\label{eq:fam2}
 \rho_{Y^1}(k)=\frac{\|\hat w-w^*\|^2}{L^*}\ \ge\ c_1\Big(\frac sn-p\Big)^2,
 \qquad c_1=\frac{c^4}{d\kappa\,p\,(1-pd\kappa)} .
\end{equation}

\emph{Conclusion.} Let $M\ge n+2$, let $Q$ be a law on selections of
budget $n$, and put $\tau=s(k)/n$ and $v=\E_Q(\tau-p)^2$. Since
$pd\kappa+(1-p)c^2=1$, we have
$\tau d\kappa+(1-\tau)c^2=1+(\tau-p)(d\kappa-c^2)$, and therefore
\[
 f(\tau):=\frac{d\kappa\,\tau}{p\,(\tau d\kappa+(1-\tau)c^2)^2}\ \ge\ d\kappa\,\big(1-\ell\,|\tau-p|\big),
 \qquad \ell=\max\{1/p,\ 2(d\kappa-c^2)\},
\]
for all $\tau\in[0,1]$: for $\tau\le p$ the denominator is at most $p$ and
$f(\tau)\ge d\kappa\,\tau/p$, and for $\tau\ge p$ we use $\tau/p\ge1$ and
$(1+x)^{-2}\ge1-2x$ for $x\ge0$. By \eqref{eq:fam1}, \eqref{eq:fam2} and
Jensen's inequality, and because a supremum over labels is at least an
average,
\[
 n\,\sup_Y\E_Q\rho_{(X,Y)}(k)\ \ge\ \max\Big\{\Big(1-\frac n{M-1}\Big)d\kappa\,\big(1-\ell\sqrt v\big),\ n\,c_1\,v\Big\}.
\]
If $v\le n^{-2/3}$ the first term is at least
$(1-\frac n{M-1})\,d\kappa\,(1-\ell\,n^{-1/3})$, and otherwise the second is
at least $c_1n^{1/3}$. The constants $\ell$ and $c_1$ depend on $d$,
$\kappa$ and $p$ only. Taking the supremum over the admissible $M$, which
is part of the supremum over designs,
\[
 n\,B_{d,\kappa}(n)\ \ge\ \min\big\{d\kappa\,(1-\ell\,n^{-1/3}),\ c_1\,n^{1/3}\big\},
\]
and hence $\liminf_nn\,B_{d,\kappa}(n)\ge d\kappa$.

\section{Rounding bounds in Theorems~\ref{thm:one} and~\ref{thm:several}}\label{app:aware}

This appendix proves the upper bounds $\Gamma_{d,m}$ and $\Lambda_d$, the case $d=1$ of Theorem~\ref{thm:several}, and the lower bound $\frac12$ of Theorem~\ref{thm:one}.

\subsection{The upper bound \texorpdfstring{$d\kappa\,\Gamma_{d,m}$}{with rank-one rounding}}

We prove $\limsup_nn\,A^{(m)}_{d,\kappa}(n)\le d\kappa\,\Gamma_{d,m}$ for all $d,m\ge1$. This is the upper bound of Theorem~\ref{thm:several}, and for $m=1$ the bound $d\kappa\Lambda_d$ of Theorem~\ref{thm:one}. Put $R=d\kappa$. Let $D$ be a dataset with $\kappa_D\le\kappa$ and $L^*_D>0$. Whiten it, subtract its full fit, and write $e_i$ for the residuals, so that
\[
 \tfrac1N\sum_ix_ix_i^\top=I,\qquad\tfrac1N\sum_ix_ie_i^\top=0,\qquad\tfrac1N\sum_i\|e_i\|^2=L^*,\qquad\|x_i\|^2\le R .
\]
By Carath\'eodory's theorem these moments are reproduced by a probability vector $p$ on at most $T=d(d+1)/2+dm+2$ of the rows. Let
\[
 a_i=np_i-\lfloor np_i\rfloor,\qquad H_i=\frac{x_ie_i^\top}{\sqrt{nRL^*}},\qquad\text{so that}\qquad
 \sum_ia_i\|H_i\|_F^2\le\sum_inp_i\|H_i\|_F^2\le1 .
\]
The $H_i$ have rank at most one. Fix $M_0$ and $\delta>0$. Give the rounding increment $k_i=0$ to the rows with $\|H_i\|_F>M_0$; since $a_i\|H_i\|_F\le1/M_0$ for these, their total target has norm at most $T/M_0$. For each number of remaining rows, which is at most $T$, the set of parameters with $\operatorname{rank}H_i\le1$, $\|H_i\|_F\le M_0$, $0\le a_i\le1$ and $\sum a_i\|H_i\|_F^2\le1$ is compact. At points of zero energy take all counts zero; at all other points the definition of $\Gamma_{d,m}$ gives counts with error less than $\sqrt{\Gamma_{d,m}}+\delta$; by continuity these counts still work in a neighborhood, and a finite subcover gives a bound $K=K(d,m,M_0,\delta)$ such that some counts $k_i\le K$ satisfy
\[
 \Big\|\sum_i(k_i-a_i)H_i\Big\|_F\le\sqrt{\Gamma_{d,m}}+\delta+T/M_0 .
\]
Select $\lfloor np_i\rfloor+k_i$ copies of row $i$. The sum of these numbers differs from $n$ by at most $T(K+1)$. Correct it at a row $i_0$ with $p_{i_0}\ge1/T$, which is possible for large $n$; since $n\,p_{i_0}\|H_{i_0}\|_F^2\le1$, we have $\|H_{i_0}\|_F\le\sqrt{T/n}$, and the correction changes the error by $O(n^{-1/2})$. All counts differ from $np_i$ by bounded amounts, so the selected design is $nI+O(1)$, uniformly over datasets. The selected gradient is $\sum_i(k_i-a_i)x_ie_i^\top+O(1)\sqrt{L^*}$ because $\sum_ip_ix_ie_i^\top=0$, and \eqref{eq:defect} gives
\[
 n\,\rho_D\le R\big(\sqrt{\Gamma_{d,m}}+\delta+T/M_0\big)^2+o(1)
\]
uniformly over datasets. Let $n\to\infty$ and then $M_0\to\infty$, $\delta\to0$.

\subsection{Lower bound for one feature}

Let $d=1$. By Proposition~\ref{prop:Lambda}(i) it suffices to show
$\liminf_nn\,A^{(m)}_{1,\kappa}(n)\ge\kappa\Lambda(\phi,a)$ for linearly
independent $\phi_1,\dots,\phi_q\in\R^m$ and rational $a\in(0,1)^q$. Let $s=\sum a_i$, $r=\sum a_i\phi_i$,
$C=\sum a_i\|\phi_i\|^2$. For large $n$ consider the dataset with rare rows
$(\sqrt\kappa,\sqrt n\,\phi_i)$ of mass $a_i/n$ and one frequent row
$(c,y_0)$ of mass $1-s/n$, where $c^2=(1-s\kappa/n)/(1-s/n)\in(0,1]$ and
$y_0={-\sqrt{\kappa}\,r/(\sqrt n\,c\,(1-s/n))}$. Then $N^{-1}\sum x_i^2=1$,
the normalized leverage is $\kappa$, $W^*=0$ and $L^*=C+O(1/n)$. A selection
with $k_i$ copies of the $i$th rare row, $K=\sum k_i$, and $n-K$ frequent
rows has a positive design and
\[
 \hat W=\frac{\sqrt{\kappa n}\,\big[\sum_ik_i\phi_i-\frac{n-K}{n-s}\,r\big]}{\kappa K+(n-K)c^2},
 \qquad \sum_ik_i\phi_i-\frac{n-K}{n-s}\,r=\sum_i(k_i-a_i)\phi_i+\frac{K-s}{n-s}\,r .
\]
Consider, along a subsequence that attains $\liminf_nn\min_k\rho(k)$, which we may assume to be finite, optimal selections; for them $n\|\hat W\|^2$ is bounded. The denominator is at most $\kappa n$, so the bracket is $O(1)$, hence $\sum_ik_i\phi_i=O(1)$, and by linear independence the $k_i$ are bounded. After passing to a further subsequence they are constant. Then the denominator is $n(1+o(1))$ and the bracket tends to $\sum_i(k_i-a_i)\phi_i$, whose squared norm is at least $\Lambda(\phi,a)C$. Hence $\liminf_nn\min_k\rho(k)\ge\kappa\Lambda(\phi,a)$,
which proves the case $d=1$ of Theorem~\ref{thm:several} and, since $\Lambda_1=\frac12$, the lower bound of Theorem~\ref{thm:one} for~$d=1$.

\subsection{The lower bound \texorpdfstring{$d\kappa/2$}{one half}}

We prove the lower bound of Theorem~\ref{thm:one} for $d\ge2$. Let $m=1$. In this subsection we use the
coordinates in which $\Sigma=I/d$. Let $n>d\kappa$ and $p=1/(2n)$. Put
$\ell=(1-d\kappa p)/(d(1-p))$ and $h=(d-1)/(d(1-p))$. Give the row
$(\sqrt\kappa\,\be_1,\sqrt\kappa)$ mass $p$ and each of the $2(d-1)$ rows
$(\sqrt\ell\,\be_1\pm\sqrt h\,\be_j,0)$, $2\le j\le d$, mass $(1-p)/(2(d-1))$.
Then $\Sigma=I/d$, the number $x^\top\Sigma^{-1}x/d$ is $\kappa$ for the first row and
$(1-\kappa p)/(1-p)\le1\le\kappa$ for the others, the full fit is
$w^*=d\kappa p\,\be_1$ and $L^*=\kappa p(1-d\kappa p)$.

A selection without the first row has zero labels and minimum-norm fit $0$,
with excess loss $\|w^*\|^2/d$ and excess ratio $d\kappa p/(1-d\kappa p)=\frac{d\kappa}{2n}(1+O(1/n))$.
Consider a selection with $t\ge1$ copies of the first row and $k_j^\pm$
copies of the rows $\sqrt\ell\,\be_1\pm\sqrt h\,\be_j$. Minimizing the selected
loss over $w_j$ for each $j\ge2$ leaves the profile loss
$t\kappa(w_1-1)^2+\vartheta w_1^2$ with
$\vartheta=4\ell\sum_{j\ge2}k_j^+k_j^-/(k_j^++k_j^-)\le(n-t)\ell$ (terms with
$k_j^++k_j^-=0$ are omitted). Hence every least-squares minimizer has
\[
 w_1=\frac{t\kappa}{t\kappa+\vartheta}\ \ge\ \frac{t\kappa}{t\kappa+(n-t)\ell}\ \ge\ \frac{\kappa}{\kappa+(n-1)\ell}=\frac{d\kappa}n\Big(1+O\Big(\frac1n\Big)\Big).
\]
Since $w_1^*=d\kappa/(2n)$ and $L^*=\frac\kappa{2n}(1+O(1/n))$, the excess ratio is at least $(w_1-w_1^*)^2/(dL^*)\ge\frac{d\kappa}{2n}(1+O(1/n))$. The constants in the $O(1/n)$ depend on $d$ and $\kappa$ only. Hence $\underline\alpha^{(1)}_{d,\kappa}\ge\frac12$.

\section{Proofs for one label}\label{app:one}

This appendix proves Theorem~\ref{thm:U} and deduces the upper bound $c_*$ of Theorem~\ref{thm:one}. We use the notation of Section~\ref{sec:game}. Averages over the finite type space $\mathcal X$ are written $\av{\cdot}=\E_\nu[\cdot]$, with larger brackets around long expressions, expectations and probabilities over random counts are written $\E$ and $\Prob$, and $Z=|\delta|^2$. Norms of vectors are written $|\cdot|$, operator norms $\|\cdot\|$, and inner products $\langle\cdot,\cdot\rangle$ or $x\cdot y$. In this appendix $n_x$ and $\ell_x$ denote the scalar functions of Section~\ref{sec:game}; the budget $n$ appears only in Section~\ref{app:transfer}. An \emph{atom} is a rare gradient $g_i$ with its rate $a_i$. The symbols $\rho$, $Q$, $D$ and $m_x$ defined below are used until the end of the proof of Theorem~\ref{thm:menu}; in Section~\ref{app:transfer} and in Appendix~\ref{app:algo}, $D$ is the dataset and $\rho$ the excess ratio. Put
\[
 \rho(v)=\min(1,|v|),\qquad\sigma_x^2=x^\top Sx,\qquad
 \omega_x=\begin{cases}\min(1,|n_x|/|\ell_x|),&\ell_x\ne0,\\1,&\ell_x=0.\end{cases}
\]

\subsection{Four strategies}\label{app:strategies}

Throughout this appendix
\begin{equation}\label{eq:params}
 \begin{gathered}
 J=10^{2364},\qquad\varepsilon=J^{-1},\qquad\theta=5\cdot10^{-24},\qquad\varrho=2\cdot10^{-33},\\
 t=10^{-15},\qquad\cutoff=6\cdot10^7,\qquad\gamma=\tfrac9{2000}.
 \end{gathered}
\end{equation}
For a type with $\sigma_x>0$ let
\[
 I_x=\{i:|\langle x,g_i\rangle|>\sigma_x/\sqrt\varrho\},\qquad p_x=\sum_{i\in I_x}a_i,\qquad
 m_x=\min\Big\{p_x,\ \frac3{|s_x|}\sum_{i\in I_x}a_i|\langle x,g_i\rangle|\Big\},
\]
with $m_x=p_x$ if $s_x=0$; if $\sigma_x=0$, then $I_x=\varnothing$ and $p_x=m_x=0$. Since $\sigma_x^2=\sum_ia_i\langle x,g_i\rangle^2$,
\begin{equation}\label{eq:psmall}
 m_x\le p_x\le\varrho\qquad\text{for all }x .
\end{equation}
 Indices in $I_x$ are called \emph{large} at $x$. Let
\[
 B_i=\{x:i\in I_x,\ |s_x+\langle x,g_i\rangle|\le\gamma|\langle x,g_i\rangle|\},\qquad p'_x=\sum_{i:x\in B_i}a_i\le p_x .
\]
The strategies are the following. Every Poisson law in them is conditioned on $\sum_ik_i\le\overline K:=\lceil J(1+A)\rceil$: in (S1) the two draws are independent draws from the conditioned law, and in (S4) the joint law of $(N_i,M_i)_i$ is conditioned. We put $\sgn0=0$.
\begin{enumerate}
\item[(S1)] The draw of smaller cost among two independent draws of $k$ with independent coordinates $k_i\sim\Pois(a_i)$, and $h=0$.
\item[(S2)] Independent $k_i\sim\Pois(a_i)$ and the density of Lemma~\ref{lem:gain}.
\item[(S3)] Independent $k_i\sim\Pois(a_i)$. Let $\mathcal E_x$ be the event that $\sum_{I_x}k_i=1$ and $|n_x-g|\le|g|/2$, where $g=\langle x,g_i\rangle$ is the scalar projection of the only large arrival $i$. Put
\[
 \beta_x=\omega_x\one_{\mathcal E_x},\qquad\bar m=\av m,\qquad\mu_k=\av{\sgn(n\ell)\beta}.
\]
If $\bar m=0$, then $h=0$. Otherwise put $\lambda_k=\av\beta/\bar m$, and
\[
 h_x=t\Bigl(\sgn(n_x\ell_x)\beta_x-\frac{\mu_k}{\bar m}\,m_x\Bigr)\quad\text{if }\lambda_k\le\cutoff,\qquad h=0\quad\text{if }\lambda_k>\cutoff .
\]
\item[(S4)] Independent $N_i\sim\Pois(a_i/2)$ and $M_i\sim\Pois(a_i/4)$, and $k_i=N_i+2M_i$. Let $D={\{x:\sum_{i:x\in B_i}M_i\ge1\}}$ and $m_D=\av{\one_D}$. If $\av{p'}>0$ and $m_D\le16\av{p'}$, then
\[
 h=4\one_D-\lambda p'\one_{D^c},\qquad\lambda=\frac{4m_D}{\av{p'\one_{D^c}}},
\]
and $h=0$ otherwise.
\end{enumerate}

\begin{theorem}\label{thm:menu}
For every affine frame and every rare configuration with $C>0$ at least one of the strategies (S1) to (S4) has mean cost at most $(1-J^{-1})\,C$. In all of them $\sum_ik_i\le\lceil J(1+A)\rceil$, $\av h=0$ and $-\frac12\le h\le4$.
\end{theorem}

This implies Theorem~\ref{thm:U}: some outcome costs at most the mean, and if $C=0$, then all $g_i$ vanish and $k=0$, $h=0$ have cost zero.

\emph{Admissibility.} For (S1) it is trivial, and for (S2) it is part of Lemma~\ref{lem:gain}. In (S3), $\av h={t(\mu_k-\mu_k)}=0$, and $|\mu_k|/\bar m\le\lambda_k\le\cutoff$ and \eqref{eq:psmall} give $|h|\le t(1+\cutoff\varrho)\le2t$. In (S4), $\av h=4m_D-4m_D=0$; moreover $\av{p'\one_{D^c}}\ge\av{p'}-\varrho m_D\ge\av{p'}/2$, so that $0\le\lambda\le128$ and $-128\varrho\le h\le4$.

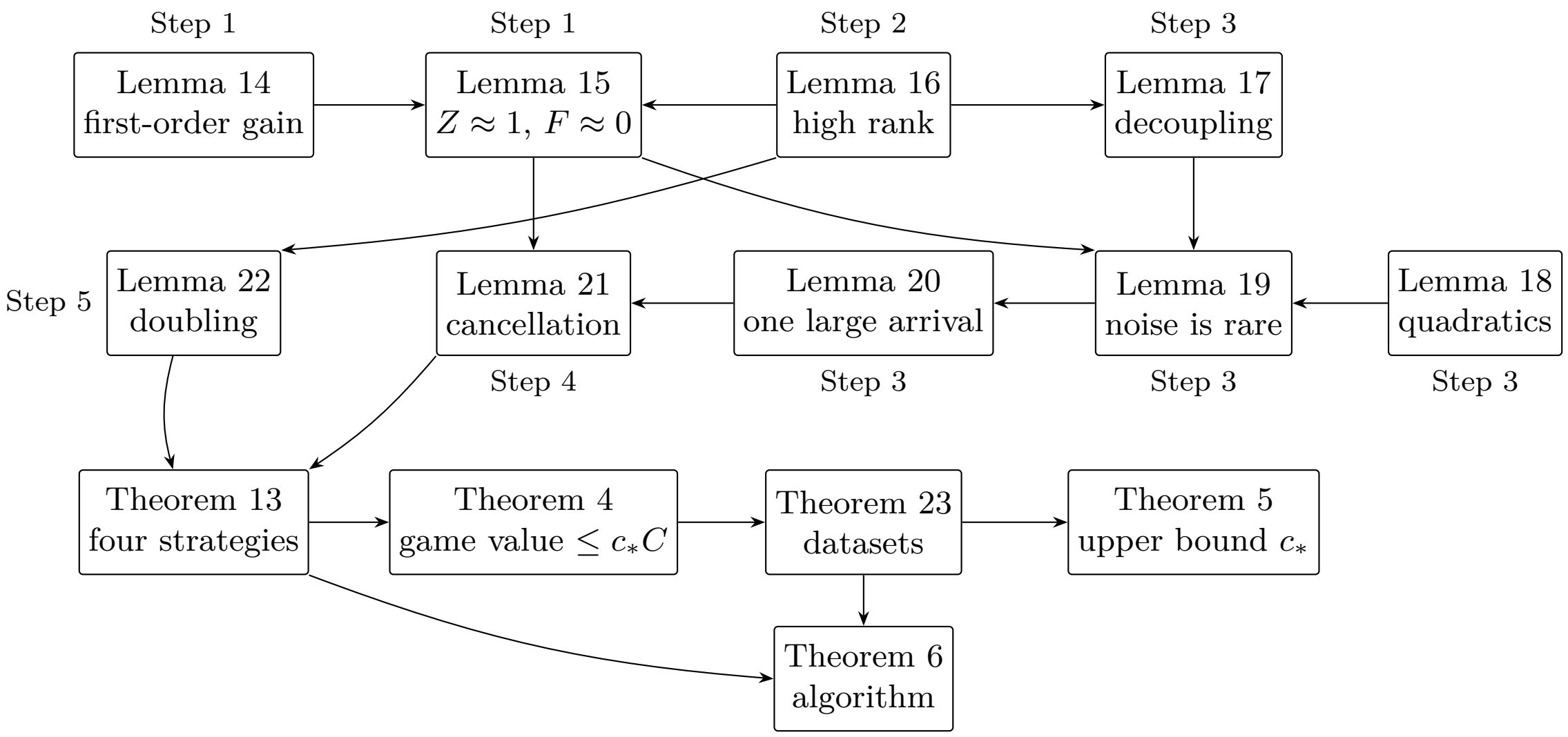
\begin{figure}[t]
\centering
\begin{tikzpicture}[font=\footnotesize,x=0.93cm,y=1cm,
 box/.style={draw,rounded corners=1pt,align=center,inner sep=2.5pt,minimum height=10mm},
 >={Stealth[length=4pt]}]
\node[box] (g) at (0,0) {Lemma~\ref{lem:gain}\\first-order gain};
\node[box] (s) at (3.5,0) {Lemma~\ref{lem:shell}\\$Z\approx1$, $F\approx0$};
\node[box] (r) at (6.9,0) {Lemma~\ref{lem:rank}\\high rank};
\node[box] (c) at (10.3,0) {Lemma~\ref{lem:conc}\\decoupling};
\node[box] (w) at (13.2,-1.9) {Lemma~\ref{lem:three}\\quadratics};
\node[box] (a) at (10.3,-1.9) {Lemma~\ref{lem:rare}\\noise is rare};
\node[box] (o) at (6.9,-1.9) {Lemma~\ref{lem:arrival}\\one large arrival};
\node[box] (k) at (3.5,-1.9) {Lemma~\ref{lem:cancel}\\cancellation};
\node[box] (b) at (0,-1.9) {Lemma~\ref{lem:double}\\doubling};
\foreach \v/\t in {g/{Step 1},s/{Step 1},r/{Step 2},c/{Step 3}} \node[anchor=south,font=\scriptsize] at (\v.north) {\t};
\foreach \v/\t in {w/{Step 3},a/{Step 3},o/{Step 3},k/{Step 4}} \node[anchor=north,font=\scriptsize] at (\v.south) {\t};
\node[anchor=east,font=\scriptsize] at (b.west) {Step 5};
\node[box] (n) at (0,-4) {Theorem~\ref{thm:menu}\\four strategies};
\node[box] (u) at (3.5,-4) {Theorem~\ref{thm:U}\\game value $\le c_*C$};
\node[box] (t) at (6.9,-4) {Theorem~\ref{thm:data}\\datasets};
\node[box] (f) at (10.3,-4) {Theorem~\ref{thm:one}\\upper bound $c_*$};
\node[box] (m) at (6.9,-5.5) {Theorem~\ref{thm:algo}\\algorithm};
\draw[->] (g)--(s); \draw[->] (r)--(s); \draw[->] (r)--(c); \draw[->] (c)--(a); \draw[->] (w)--(a);
\draw[->] (a)--(o); \draw[->] (o)--(k); \draw[->] (s)--(k);
\draw[->] (s.south east) to[bend right=8] (a.north west);
\draw[->] ([xshift=-2mm]b.south) to[bend right=15] ([xshift=-2mm]n.north); \draw[->] (k.south west) to[bend left=8] (n.north east);
\draw[->] (r.south west) to[bend left=6] (b.north east);
\draw[->] (n)--(u); \draw[->] (u)--(t); \draw[->] (t)--(f);
\draw[->] (n.south east) to[bend right=8] (m.west); \draw[->] (t)--(m);
\end{tikzpicture}
\caption{Principal dependencies in the proof for one label. Lemmas~\ref{lem:shell} and~\ref{lem:rare} to~\ref{lem:cancel} assume that all four strategies fail. The other lemmas do not; Lemma~\ref{lem:shell} uses Lemma~\ref{lem:rank} only for $\tau>0$, and the bound of Lemma~\ref{lem:double} contradicts \eqref{eq:lower}. Theorem~\ref{thm:algo} uses the four strategies of Theorem~\ref{thm:menu} and the construction in the proof of Theorem~\ref{thm:data}.}
\label{fig:map}
\end{figure}

For the proof of Theorem~\ref{thm:menu} we may scale $g_i$ and $\phi$ and assume
\begin{equation}\label{eq:hyp}
 C=1,\quad\text{and all four strategies have mean cost larger than }1-\varepsilon .
\end{equation}
We derive a contradiction. From now on $k$ denotes one \emph{unconditioned} draw with independent coordinates $k_i\sim\Pois(a_i)$ unless another law is specified; in particular it is not the better draw of (S1). Then $\E Z=C=1$, and the event $Q=\{\sum_ik_i\le\overline K\}$ has probability at least ${1-\varepsilon}$ by Markov's inequality; the same holds for the law of (S4), under which $\E k_i=a_i$ as well. Figure~\ref{fig:map} shows the structure of the argument.

\subsection{The first-order gain and the shell}

\begin{lemma}\label{lem:gain}
For every $k$, put $F=F(k)=\min_{c\in\R}\av{\omega|n\ell-c|}$. There is an $h$ with $\av h=0$, $|h|\le2/9$, and
\begin{equation}\label{eq:gain}
 |w-r|^2\le Z-\frac29\frac{F^2}{Z},
\end{equation}
where the fraction is zero when $Z=0$.
\end{lemma}
\begin{proof}
Weighted medians give
$F=\max\{\av{\psi n\ell}:|\psi|\le\omega,\ \av\psi=0\}$.
Indeed, subtracting any $c$ gives the upper bound, and weighted signs
about a median attain it, with a fractional sign on ties. This also
covers zero total weight. In particular, $0\le F\le\av{n^2}=Z$.
Recall \eqref{eq:fitmain}: ${w-r}={\delta-(I+H)^{-1}b}$ with
$H=\av{hxx^\top}$ and $b=\av{h\ell x}$. Isotropy gives $|\av{ux}|^2\le\av{|u|^2}$ and
$\|H\|\le\|h\|_\infty$. Choose a maximizing $\psi$ and $h=\xi\psi$ with $0\le\xi<1$.
Then $|b|\le\xi\sqrt Z$ and $\delta\cdot b=\xi F$, and~hence
\[
 |w-r|^2\le Z-2\xi F+
 \left(\frac{2\xi^2}{1-\xi}+\frac{\xi^2}{(1-\xi)^2}\right)Z.
\]
The two terms in parentheses bound the inverse correction in the cross term and
the squared correction, respectively. Choose $\xi=2F/(9Z)$, or $\xi=0$
if $Z=0$. On $[0,2/9]$ the coefficient
$(3-2\xi)/(1-\xi)^2$ is increasing and at most $207/49<9/2$;
this proves \eqref{eq:gain}.
\end{proof}

\begin{lemma}\label{lem:shell}
Assume \eqref{eq:hyp}. Then
\begin{equation}\label{eq:shell}
 0<\tau:=\E|Z-1|\le11\varepsilon,\qquad\E F\le5\sqrt\varepsilon,
\end{equation}
and the costs of (S3) and (S4), for the unconditioned laws, satisfy
\begin{equation}\label{eq:lower}
 \E|w-r|^2\ge1-2\varepsilon .
\end{equation}
\end{lemma}
\begin{proof}
Let $Z_1,Z_2$ be independent with the law of $Z$ given $Q$, and put $\bar Z=\E[Z\mid Q]$. Then $\bar Z\le1/\Prob(Q)\le1+2\varepsilon$, and (S1) gives $\E\min(Z_1,Z_2)>1-\varepsilon$, so that $\bar Z>1-\varepsilon$ and
\[
 \E|Z_1-\bar Z|\le\E|Z_1-Z_2|=2\big(\bar Z-\E\min(Z_1,Z_2)\big)\le6\varepsilon,\qquad
 \E\big[|Z-1|\,\big|\,Q\big]\le8\varepsilon .
\]
Moreover $\E[Z;Q^c]=1-\Prob(Q)\bar Z\le1-(1-\varepsilon)^2\le2\varepsilon$. Hence $\tau\le8\varepsilon+\Prob(Q^c)+\E[Z;Q^c]\le11\varepsilon$. By (S2) and \eqref{eq:gain}, $\frac29\,\E[F^2/Z\mid Q]\le\bar Z-1+\varepsilon\le3\varepsilon$, and the Cauchy--Schwarz inequality gives $\E[F;Q]\le(\frac{27}2\varepsilon(1+2\varepsilon))^{1/2}\le4\sqrt\varepsilon$, while $\E[F;Q^c]\le\E[Z;Q^c]\le2\varepsilon$. The costs are nonnegative, so in (S3) and (S4) the unconditioned mean is at least $\Prob(Q)(1-\varepsilon)\ge1-2\varepsilon$. That $\tau>0$ will follow from \eqref{eq:rank} below: $\tau=0$ would give $\tr S^2=0$, but $\tr S=C=1$.
\end{proof}

\subsection{High effective rank and decoupling}

\begin{lemma}\label{lem:rank}
Let $C=1$ and $\tau=\E|Z-1|$. Then
\begin{equation}\label{eq:rank}
 \tr S^2\le14\,\tau .
\end{equation}
If $\tau>0$ and the atoms with $|g_i|^2\ge16\,\tau$ are called \emph{omitted} and the others \emph{retained}, then
\begin{equation}\label{eq:diffuse}
 E_{\mathrm{om}}:=\sum_{\mathrm{omitted}}a_i|g_i|^2\le4\,\tau,\qquad\sum_{\mathrm{retained}}a_i|g_i|^3\le4\sqrt\tau .
\end{equation}
\end{lemma}
\begin{proof}
Thin any atoms to rates $b_i\le a_i$ with $R_b=\sum b_i\le\frac12$,
energy $C_b=\sum b_i|g_i|^2$, and mean $r_b=\sum b_i g_i$.
Conditional on no thinned arrivals, $\E Z=1-C_b+|r_b|^2$, and conditioning on this event increases the expectation of $|Z-1|$ by at most the factor $e^{R_b}$.
Since $|r_b|^2\le R_bC_b$,
\[
 (1-R_b)C_b\le C_b-|r_b|^2\le e^{R_b}\tau,\qquad C_b\le2e^{1/2}\tau<4\,\tau\quad\text{if }\tau>0.
\]
If the omitted atoms had total rate at least $\frac12$, thinning them to total rate $\frac12$ would give $C_b\ge8\,\tau$, a contradiction. So their total rate is less than $\frac12$, and the bound applies to all of them. Retained atoms have $|g_i|<4\sqrt\tau$, and $\sum a_i|g_i|^2=1$. This proves \eqref{eq:diffuse}. For \eqref{eq:rank} split $\delta=X+Y$ into independent centered
Poisson halves. The conditional Jensen inequality gives
$\E\big||X|^2-\frac12\big|\le\tau$ and $\E\big||Y|^2-\frac12\big|\le\tau$,
so $\E|X\cdot Y|\le3\tau/2$. Where both squared norms are at most
one, $(X\cdot Y)^2\le|X\cdot Y|$. On the complement, independence
and $\E[|X|^2;|X|^2>1]\le2\tau$ bound the expectation of $|X|^2|Y|^2$ by
$2\tau$. Thus $\tr S^2=4\,\E(X\cdot Y)^2\le14\,\tau$.
\end{proof}

Put $d\mu=\sigma_x^2d\nu$, $Y_x=n_x/\sigma_x$, and $z_i(x)=\langle x,g_i\rangle/\sigma_x$.
Divided expressions are used only where $\sigma_x>0$; all atom
projections vanish elsewhere. Isotropy gives
\begin{equation}\label{eq:scalar}
 \mu(\mathcal X)=1,\quad \E Y_x=0,\quad\E Y_x^2=1,\quad
 \sum_i a_i z_i(x)^2=1,\quad \int z_i z_l\,d\mu=g_i\cdot g_l.
\end{equation}

\begin{lemma}[Decoupling]\label{lem:conc}
Let $C=1$, $0<\tau=\E|Z-1|\le1$, and let $(K_{xy})$ be a deterministic family of functions on $\R^2$ with $|K_{xy}|\le1$ that are $L$-Lipschitz in each coordinate. Let $Y^{(1)}$ and $Y^{(2)}$ be independent copies of the family $(Y_x)_{x\in\mathcal X}$. Then
\begin{equation}\label{eq:factor}
 \iint\big|\E K_{xy}(Y_x,Y_y)-\E K_{xy}(Y^{(1)}_x,Y^{(2)}_y)\big|\,d\mu(x)d\mu(y)\le15\,L\,\tau^{1/6}.
\end{equation}
\end{lemma}
\begin{proof}
\emph{Smooth functions.} Suppose first that $K_{xy}$ is smooth, with bounded derivatives, and $|\partial_{uv}K_{xy}|\le L_{11}$, $|\partial_{uuv}K_{xy}|\le L_{21}$ and $|\partial_{uvv}K_{xy}|\le L_{12}$. For $\upsilon\in[0,1]$ let the arrivals of the atom $i$ be of three independent kinds: \emph{shared} ones of rate $\upsilon a_i$, which count for both coordinates, and arrivals of rate $(1-\upsilon)a_i$ for each coordinate separately. The marginal laws do not depend on $\upsilon$; at $\upsilon=0$ the coordinates are independent, and at $\upsilon=1$ the pair has the law of $(Y_x,Y_y)$. The derivative of $\E K_{xy}$ in $\upsilon$ is the expectation of
\[
 \sum_ia_i\big[K(u+z_i(x),v+z_i(y))-K(u+z_i(x),v)-K(u,v+z_i(y))+K(u,v)\big]
\]
at the random point $(u,v)$, because the centering does not depend on $\upsilon$. For omitted atoms the bracket is at most $L_{11}|z_i(x)z_i(y)|$, and $\int|z_i|\,d\mu\le|g_i|$ by \eqref{eq:scalar}; their contribution to the left-hand side of \eqref{eq:factor} is at most $L_{11}E_{\mathrm{om}}$. For retained atoms the bracket differs from $z_i(x)z_i(y)\,\partial_{uv}K(u,v)$ by at most $\frac12L_{21}z_i(x)^2|z_i(y)|+\frac12L_{12}|z_i(x)|z_i(y)^2$, which contributes at most $\frac12(L_{21}+L_{12})\sum_{\mathrm{retained}}a_i|g_i|^3$. The remaining coefficient $c_{xy}=\sum_{\mathrm{retained}}a_iz_i(x)z_i(y)$ satisfies $\iint c_{xy}^2\,d\mu d\mu=\sum_{i,l\ \mathrm{retained}}a_ia_l{(g_i\cdot g_l)^2}\le\tr S^2$. By Lemma~\ref{lem:rank} the left-hand side of \eqref{eq:factor} is at most
\[
 L_{11}\big(\sqrt{14\,\tau}+4\,\tau\big)+2\,(L_{21}+L_{12})\sqrt\tau .
\]

\emph{Lipschitz functions.} Let $\varsigma=\tau^{1/6}$ and convolve $K_{xy}$ in each coordinate with the law of the sum of two independent uniform variables on $[-\varsigma,\varsigma]$, and with an arbitrarily narrow smooth kernel. The change of $K_{xy}$ is at most $\frac43L\varsigma$, up to an arbitrarily small error. One derivative of the new function is bounded by $L$, and every further derivative can be put on a uniform factor, which costs a factor $1/\varsigma$, since $\partial_u(f*U_\varsigma)(u)=(f(u+\varsigma)-f(u-\varsigma))/(2\varsigma)$ for the uniform density $U_\varsigma$. Thus $L_{11}\le L/\varsigma$ and $L_{21},L_{12}\le L/\varsigma^2$, and the left-hand side of \eqref{eq:factor} is at most
\[
 L\Big[\frac{\sqrt{14\,\tau}+4\,\tau}\varsigma+\frac{4\sqrt\tau}{\varsigma^2}+\frac83\varsigma\Big]
 \le L\Big[\sqrt{14}+4+4+\frac83\Big]\tau^{1/6}\le15\,L\,\tau^{1/6}.
\]
This proves \eqref{eq:factor}.
\end{proof}

\subsection{The noise is rare}

The next lemma replaces, in a quantitative way, the fact that a nondegenerate infinitely divisible law is not carried by the roots of a quadratic equation. In this paragraph and in Lemma~\ref{lem:three}, $\theta$ may be any number in $(0,1]$. Put $D_\theta=36/\theta$ and let $\chi=\chi_\theta$ be the continuous piecewise linear even function that vanishes for $|y|\le\theta/48$ and for $|y|\ge D_\theta+1$, equals one for $\theta/24\le|y|\le D_\theta$, and is linear in between. It is $(48/\theta)$-Lipschitz.

\begin{lemma}\label{lem:three}
Let $0<\theta\le1$ and let $Y=W_1+W_2+W_3$ with independent identically distributed centered $W_j$ of variance at most $\frac13$. Suppose that $\Prob(|Y|\ge\theta)\ge\theta^2/16$. Then $\E\chi(Y)\ge\theta^2/32$ and
\begin{equation*}
 \E\,\chi(Y)\,\rho(\alpha Y^2+\beta Y-z)\ge10^{-13}\,\theta^9
\end{equation*}
for all $z\in\R$, $0\le\alpha\le1$ and $|\beta|\le1$ with $\max(\alpha,|\beta|)=1$.
\end{lemma}
\begin{proof}
Let $W,W'$ be independent copies of $W_1$ and $b=\theta/12$. We claim that
\[
 q_0:=\Prob(|W-W'|\ge b)\ge\theta^2/48 .
\]
Suppose not. Averaging over $W'$ gives $u\in\R$ with $p:=\Prob(|W-u|\ge b)\le q_0<\frac12$. Since $\E W=0$ and $\E W^2\le\frac13$, $|u|\le b(1-p)+|u|p+\sqrt{p/3}$, and hence $|u|\le b+2\sqrt{q_0/3}$. If all $|W_j-u|<b$, then $|Y|<3|u|+3b\le\frac\theta2+2\sqrt{3q_0}<\theta$. Therefore $\Prob(|Y|\ge\theta)\le3q_0<\theta^2/16$, a contradiction.

With $\overline H=12/\theta$, symmetry and Chebyshev's inequality give
\[
 q:=\Prob\big(W'-W\ge b,\ |W|\le\overline H,\ |W'|\le\overline H\big)\ge\frac{\theta^2}{96}-\frac2{3\overline H^2}=\theta^2\Big(\frac1{96}-\frac1{216}\Big)\ge\frac{\theta^2}{192}.
\]
Take three independent pairs $(U_j,V_j)$ of independent copies of $W$. With probability $q^3$ they all satisfy $V_j-U_j\ge b$ and $|U_j|,|V_j|\le\overline H$. On this event the sums
\[
 T_0=U_1+U_2+U_3,\quad T_1=V_1+U_2+U_3,\quad T_2=V_1+V_2+U_3,\quad T_3=V_1+V_2+V_3
\]
increase with gaps at least $b$ and lie in $[-D_\theta,D_\theta]$. At most one of them has absolute value less than $b/2=\theta/24$. Let $t_1<t_2<t_3$ be three of the others, $P(y)=\alpha y^2+\beta y$ and $\eta_0=\theta^3/50\,000$. If $|P(t_j)-z|<\eta_0$ for $j=1,2,3$, then divided differences give
\[
 |\alpha|\le\frac{2\eta_0}{b^2}=\frac{288\,\theta}{50\,000}<1,\qquad|\beta|\le\frac{2\eta_0}b+2D_\theta|\alpha|\le\frac{24+20\,736}{50\,000}<1,
\]
which contradicts $\max(\alpha,|\beta|)=1$. So some $T_j$ has $\chi(T_j)=1$ and $\rho(P(T_j)-z)\ge\eta_0$. Every $T_j$ has the law of $Y$. Hence $4\,\E\,\chi(Y)\rho(P(Y)-z)\ge\eta_0 q^3\ge\theta^9/(50\,000\cdot192^3)\ge4\cdot10^{-13}\theta^9$. Finally $\E\chi(Y)\ge\Prob(|Y|\ge\theta)-\Prob(|Y|>D_\theta)\ge\theta^2/16-\theta^2/1296\ge\theta^2/32$.
\end{proof}

\begin{lemma}\label{lem:rare}
Assume \eqref{eq:hyp}. Then
\begin{equation}\label{eq:vanish}
 \int\Prob(|Y_x|\ge\theta)\,d\mu(x)<\theta^2/8 .
\end{equation}
\end{lemma}
\begin{proof}
Suppose not. The set $\mathcal G=\{x:\Prob(|Y_x|\ge\theta)\ge\theta^2/16\}$ has $\mu$-mass at least $\theta^2/16$. For $x\in\mathcal G$, Lemma~\ref{lem:three} applies to $Y_x$: split every Poisson count into three independent ones of rates $a_i/3$. Put
\[
 t_x=s_x/\sigma_x,\qquad L_x=\max(1,|t_x|),\qquad P_x(y)=(y^2+t_xy)/L_x,\qquad q_x=\sigma_x^2L_x .
\]
The coefficients of $P_x$ are as in Lemma~\ref{lem:three}, and $n_x\ell_x=q_xP_x(Y_x)$. As $|Y_x+t_x|\le L_x(1+|Y_x|)$,
\[
 L_x\omega_x\ge\frac{|Y_x|}{1+|Y_x|}\ge\frac\theta{49}\qquad\text{where }\chi(Y_x)>0 ,
\]
and therefore, for every $c\in\R$,
\[
 \av{\omega|n\ell-c|}\ge\frac\theta{49}\int_{\mathcal G}\chi(Y_x)\,\rho\big(P_x(Y_x)-c/q_x\big)\,d\mu(x).
\]
To eliminate $c$ we compare pairs of types $x,y\in\mathcal G$. Let
\[
 K_{xy}(u,v)=\chi(u)\chi(v)\,\rho\Big(\frac{q_xP_x(u)-q_yP_y(v)}{q_x+q_y}\Big),\qquad
 I=\iint_{\mathcal G^2}K_{xy}(Y_x,Y_y)\,d\mu(x)d\mu(y).
\]
The argument of $\rho$ is $\frac{q_x}{q_x+q_y}(P_x(u)-c/q_x)-\frac{q_y}{q_x+q_y}(P_y(v)-c/q_y)$, and $\rho$ is subadditive and nondecreasing in $|\cdot|$. Hence $I$ is at most twice the last integral, for every $c$, and
\begin{equation}\label{eq:pair}
 F\ge\frac\theta{98}\,I .
\end{equation}
For independent copies, condition on the variable whose $q$ is smaller; the coefficient of the other polynomial is in $[\frac12,1]$, and $\rho(\lambda v)\ge\frac12\rho(v)$ for such $\lambda$. Lemma~\ref{lem:three} gives
\[
 \E K_{xy}(Y^{(1)}_x,Y^{(2)}_y)\ge\tfrac12\cdot10^{-13}\theta^9\cdot\tfrac1{32}\theta^2\quad(x,y\in\mathcal G),\quad
 \iint_{\mathcal G^2}\E K_{xy}(Y^{(1)}_x,Y^{(2)}_y)\,d\mu d\mu\ge\frac{\theta^{15}}{2\cdot10^{17}} .
\]
Moreover $|K_{xy}|\le1$, and $K_{xy}$ is $(123/\theta)$-Lipschitz in each coordinate: the factor $\chi$ contributes $48/\theta$, and where $\chi>0$ the polynomial contributes $2(D_\theta+1)+1\le75/\theta$. By Lemma~\ref{lem:conc}, applied with $K_{xy}=0$ outside $\mathcal G^2$, and by \eqref{eq:shell} and \eqref{eq:params},
\[
 \E I\ge\frac{\theta^{15}}{2\cdot10^{17}}-\frac{1845}\theta\,\tau^{1/6}\ge\frac{\theta^{15}}{4\cdot10^{17}},
 \qquad\text{since}\quad\tau\le11\,\varepsilon\le10^{-126}\,\theta^{96}.
\]
With \eqref{eq:pair} this gives $\E F\ge\theta^{16}/(392\cdot10^{17})>5\sqrt\varepsilon$, which contradicts \eqref{eq:shell}.
\end{proof}

\begin{lemma}\label{lem:arrival}
Assume \eqref{eq:hyp} and let $U_x=\one\{\sum_{I_x}k_i=1\}\sum_{I_x}k_i\langle x,g_i\rangle$. Then
\begin{equation*}
 \Theta:=\int\sum_ia_i\min(z_i^2,1)\,d\mu\le80\,\theta^2,\qquad
 E_0:=\E\av{(n-U)^2}\le\frac{2\Theta}\varrho+6\varrho .
\end{equation*}
\end{lemma}
\begin{proof}
For $0\le\upsilon\le1$ put $\Psi_x(\upsilon)=\sum_ia_i(1-\cos(\upsilon z_i(x)))\le \upsilon^2/2\le\frac12$. The characteristic function of $Y_x$ has modulus $e^{-\Psi_x(\upsilon)}$, so that $\Psi_x(\upsilon)\le2(1-e^{-\Psi_x(\upsilon)})\le2\,\E(1-\cos(\upsilon Y_x))$, and, by \eqref{eq:vanish}, $\int\E(1-\cos(\upsilon Y_x))\,d\mu\le \upsilon^2\theta^2/2+\theta^2/4$. Moreover
$\int_0^1(1-\cos(\upsilon z))\,d\upsilon\ge\min(z^2,1)/96$:
for $|z|\le1$ this follows from $1-\cos w\ge w^2/4$, $|w|\le1$;
for $1\le|z|\le2$ integrate only to $\frac12$; for $|z|\ge2$ use
$1-\sin z/z\ge\frac12$. Hence $\Theta\le96\cdot2\int_0^1(\upsilon^2\theta^2/2+\theta^2/4)\,d\upsilon=80\,\theta^2$.

At a type with $\sigma_x>0$ let $v=\sum_{I_x}a_i z_i^2$ and
$m^*=\sum_{I_x}a_i z_i$; then $v\le1$ and $(m^*)^2\le p_xv$.
The difference between $Y_x$ and the uncentered large-jump sum $\sum_{I_x}k_iz_i$ has second moment
${1-v+(m^*)^2}$, and the square of that sum on two or more arrivals
has expectation ${(1-e^{-p_x})v+(m^*)^2}\le2p_x$.
Thus $\E(Y_x-U_x/\sigma_x)^2\le2(1-v)+6p_x$. Finally
$\int(1-v)\,d\mu\le\Theta/\varrho$, because $z^2\le\varrho^{-1}\min(z^2,1)$ if $z^2\le\varrho^{-1}$. Types with $\sigma_x=0$ have $n_x=U_x=0$.
\end{proof}

\subsection{Cancellation}

\begin{lemma}\label{lem:cancel}
Assume \eqref{eq:hyp}. Then
\begin{equation}\label{eq:gainzero}
 \E\av{\min(n^2,|n\ell|)}\le\mathcal J:=9E_0+\frac\varepsilon t+8t+2\cutoff\sqrt{10\varrho}+20t\cutoff^2\varrho+\frac1\cutoff+\tau ,
\end{equation}
and
\begin{equation}\label{eq:cancel}
 E_1:=\Big\langle\sum_i a_i\big(\langle x,g_i\rangle+s_x\one_{B_i}(x)\big)^2\Big\rangle
 \le\frac\Theta\varrho+\frac{32E_0}{\gamma^2}+\frac{32\mathcal J}{9\gamma}+\gamma^2 .
\end{equation}
\end{lemma}
\begin{proof}
On a unique large arrival with $|n_x-g|>|g|/2$ we have $|n_x|\le3|n_x-g|$, and without a unique large arrival $U_x=0$. Hence $\E\av{n^2\one_{\mathcal E_x^c}}\le9E_0$. On $\mathcal E_x$, $\omega_x|s_x|\le\omega_x|\ell_x|+\omega_x|n_x|\le2|n_x|\le3|g|$ and $\beta_x|n_x\ell_x|=\min(n_x^2,|n_x\ell_x|)$. A unique arrival of the large index $i$ has probability $a_ie^{-p_x}$. Therefore $\E\beta_x\le m_x\le p_x\le\varrho$, and with the Cauchy--Schwarz inequality
\begin{equation}\label{eq:compbound}
 m_x|s_x|\le3\sqrt{p_x}\,\sigma_x,\qquad
 \E\av{m^2\ell^2}=\av{m^2(s^2+\sigma^2)}\le9\varrho+\varrho^2\le10\,\varrho .
\end{equation}
If $\bar m=0$, then $m=0$, hence $\beta=0$ almost surely, so that $\min(n_x^2,|n_x\ell_x|)$ vanishes on $\mathcal E_x$, and the first assertion follows from $\E\av{n^2\one_{\mathcal E_x^c}}\le9E_0$.
Otherwise put $R_k^2=\av{m^2\ell^2}$ and $T_k=\av{\beta|n\ell|}$, and consider (S3). On outcomes with $\lambda_k\le\cutoff$ we have $\|h\|_\infty\le2t$ and $\beta_x|\ell_x|\le|n_x|$, and Bessel's inequality gives
\[
 |b|\le t(\sqrt Z+\cutoff R_k),\qquad\delta\cdot b\ge tT_k-t\cutoff R_k\sqrt Z .
\]
Expanding \eqref{eq:fitmain}, using $\|(I+H)^{-1}-I\|\le2t/(1-2t)$ and $1-2t\ge\frac34$, yields
\[
 |w-r|^2\le Z-2tT_k+16t^2Z+4t\cutoff R_k\sqrt Z+4t^2\cutoff^2R_k^2 ;
\]
before enlargement the coefficients of $t^2Z$ and $t^2\cutoff^2R_k^2$ are at most $64/9$ and $16/9$, and the mixed coefficient is at most $2t+(80/9)t^2\le4t$. On the other outcomes $|w-r|^2=Z$. Taking expectations and using \eqref{eq:lower}, $\E Z=1$ and \eqref{eq:compbound},
\[
 \E[T_k;\lambda_k\le\cutoff]\le\frac\varepsilon t+8t+2\cutoff\sqrt{10\varrho}+20t\cutoff^2\varrho .
\]
Since $\E\lambda_k\le1$ and $T_k\le Z$, $\E[T_k;\lambda_k>\cutoff]\le\Prob(\lambda_k>\cutoff)+\tau\le\cutoff^{-1}+\tau$. As $\beta_x|n_x\ell_x|=\min(n_x^2,|n_x\ell_x|)$ on $\mathcal E_x$, this proves \eqref{eq:gainzero}.

For the second assertion, call a large index $i\in I_x$ with projection $g$ \emph{bad} at $x$ if $|s_x+g|>\gamma|g|$. On the event of a unique arrival of a bad index, discard the outcomes with $|n_x-g|>\frac\gamma4|g|$; the energy $g^2$ of the discarded jumps is at most $16(n_x-U_x)^2/\gamma^2$. On the rest, $|n_x|\ge\frac34|g|$ and $|\ell_x|\ge\frac{3\gamma}4|g|$, so that $\min(n_x^2,|n_x\ell_x|)\ge\frac{9\gamma}{16}g^2$. As unique arrivals have probability $a_ie^{-p_x}$ and $e^\varrho<2$, the bad indices have energy at most
\[
 \Big\langle\sum_{i\ \mathrm{bad}}a_i\langle x,g_i\rangle^2\Big\rangle\le2\Big(\frac{16E_0}{\gamma^2}+\frac{16\mathcal J}{9\gamma}\Big).
\]
The indices outside $I_x$ have energy at most $\int\sum_{i\notin I_x}a_iz_i^2\,d\mu\le\Theta/\varrho$, because $z_i^2\le\varrho^{-1}\min(z_i^2,1)$ for them. On $B_i$ the summand in \eqref{eq:cancel} is at most $\gamma^2\langle x,g_i\rangle^2$, and these energies add up to at most $\av{\sum a_i\langle x,g_i\rangle^2}=1$.
\end{proof}

\subsection{Doubling}

\begin{lemma}\label{lem:double}
Let $C=1$, let $0<\tau\le1$, where $\tau=\E|Z-1|$ for independent counts $k_i\sim\Pois(a_i)$, and suppose that $E_1\le d_0^2$ with $0\le d_0<1$. Then the cost of (S4), for the unconditioned law, satisfies
\begin{equation}\label{eq:Ubound}
 \begin{gathered}
 \E|w-r|^2\le\frac78+\frac{16}{25}\Big(\frac92d_0+\frac34\varrho+30\sqrt\tau\Big)+\frac52E_*+E_*^2,\\
 E_*=5d_0+260\sqrt\varrho\,(1+d_0)+448\,\varrho .
 \end{gathered}
\end{equation}
\end{lemma}
\begin{proof}
Except in the definition of $\tau$, all expectations are for the law of (S4). Then $\E k_i=a_i$, $\Var k_i=3a_i/2$, $\E Z=\frac32$, and we first show that
\begin{equation}\label{eq:double-shell}
 \tau_D:=\E|Z-\tfrac32|\le30\sqrt\tau .
\end{equation}
A centered compound Poisson vector $X$ of covariance $\Xi$ has
$\Var|X|^2=2\tr\Xi^2+\sum_\xi\operatorname{rate}(\xi)|\xi|^4$:
the fourth-moment expansion consists of the three covariance pairings
and the single-arrival fourth cumulant. For the retained atoms of Lemma~\ref{lem:rank} this variance is
$\frac92\bigl(\tr S_0^2+\sum_{\mathrm{retained}}a_i|g_i|^4\bigr)\le\frac92(14\tau+16\tau)=135\,\tau$, where $S_0=\sum_{\mathrm{retained}}a_ig_ig_i^\top\preceq S$. The retained vector has mean squared norm $\frac32(1-E_{\mathrm{om}})$ and the omitted one $\frac32E_{\mathrm{om}}$, so the Cauchy--Schwarz inequality gives $\tau_D\le{\sqrt{135\,\tau}+3\sqrt{E_{\mathrm{om}}}+3E_{\mathrm{om}}}\le18\sqrt\tau+12\,\tau$.

By \eqref{eq:psmall}, $p'_x\le p_x\le\varrho$. Write $p$ for $p'_x$, and $N_x=\sum_{i:x\in B_i}N_i$, $M_x=\sum_{i:x\in B_i}M_i$, $\zeta_x=N_x+2M_x$, so that $D=\{x:M_x\ge1\}$. The model $n_x^0=-s_x(\zeta_x-p_x')$ satisfies
\[
 \E\av{(n-n^0)^2}=\tfrac32E_1\le\tfrac32d_0^2,\qquad|\sqrt T-1|\le\sqrt{E_1}\le d_0\quad\text{for }T:=\av{s^2p'},
\]
the latter by the triangle inequality for the functions $\langle x,g_i\rangle$ and $-s_x\one_{B_i}$ on pairs $(x,i)$ with weights $\nu_xa_i$. For $p\in[0,1]$, subtraction of the event $M_x=0$ gives
\[
 F_0(p):=\E[(\zeta_x-p)^2;M_x\ge1]
   =\tfrac32p-e^{-p/4}(p/2+p^2/4),
\]
\[
 Q_0(p):=\E[(1-(\zeta_x-p)/2)^2;M_x\ge1]
   =1-e^{-p/4}-(p/2)e^{-p/4}+F_0(p)/4,
\]
and $|F_0(p)-p|\le3p^2/8$, $0\le Q_0(p)\le5p^2/32$, by $1-p/4\le e^{-p/4}\le1$. With Minkowski's inequality, $\sqrt{3/2}<\frac54$ and $\sqrt{5/32}<\frac25$,
\[
 \E\av{n^2\one_D}\ge1-\tfrac92d_0-\tfrac34\varrho,\qquad
 \big(\E\av{(\ell-n/2)^2\one_D}\big)^{1/2}\le q_*:=\tfrac25\sqrt\varrho\,(1+d_0)+\tfrac58d_0 .
\]
Indeed, the square root of the first quantity is at least $\sqrt{1-\frac38\varrho}\,(1-d_0)-\sqrt{3/2}\,d_0\ge1-\frac94d_0-\frac38\varrho$; if the right-hand side is positive, squaring gives the bound, and otherwise the bound is trivial.

Since $E_1<1$ we have $T>0$ and $\av{p'}>0$. Call outcomes with $m_D\le16\av{p'}$ \emph{accepted}; as $\E m_D\le\av{p'}/4$, the others have probability at most $\frac1{64}$. For every outcome put $P=\av{xx^\top\one_D}$ and $G_0=I+4P$, so that $0\preceq P\preceq I$ and $\delta^\top P\delta=\av{n^2\one_D}$. On accepted outcomes put $h^-=-\lambda p'\one_{D^c}$ and $H^-=\av{h^-xx^\top}$, and on the others $h^-=0$. Then, on accepted outcomes,
\[
 G=G_0+H^-,\qquad\|H^-\|\le128\varrho\le\tfrac12,\qquad b=2P\delta+e_b ,
\]
which defines $e_b$ there; put $e_b=0$ on the other outcomes. Write $\|f\|_{L^2(\mathrm{accepted})}$ for the norm $(\E[|f|^2;\text{accepted}])^{1/2}$. Using $\E\av{(h^-)^2\ell^2}\le128^2\av{p'^2(s^2+\frac32\sigma^2)}\le128^2(\varrho T+\frac32\varrho^2)$,
\[
 \|e_b\|_{L^2({\mathrm{accepted}})}\le4q_*+128\sqrt{\varrho\,(1+d_0)^2+\tfrac32\varrho^2}\le\tfrac{648}5\sqrt\varrho\,(1+d_0)+\tfrac52d_0+160\,\varrho .
\]
Let $u=[I-2G_0^{-1}P]\delta$ on accepted outcomes and $u=\delta$ on the others. The resolvent identity, $\|G^{-1}\|\le2$, $\|2G_0^{-1}P\|\le\frac25$ and $\E Z=\frac32$ show that the $L^2$ distance between $w-r$ and $u$, which agree on the outcomes that are not accepted, is at most
\[
 2\Big[\tfrac25\cdot128\,\varrho\,\sqrt{3/2}+\|e_b\|_{L^2(\mathrm{accepted})}\Big]\le E_* .
\]
The function $q(y)=((1+2y)/(1+4y))^2$ is convex on $[0,1]$, since $q''(y)=8(8y+5)/(1+4y)^4>0$, with $q(0)=1$ and $q(1)=\frac9{25}$. Its endpoint chord and the spectral theorem give
\[
 |[I-2G_0^{-1}P]\delta|^2\le Z-\tfrac{16}{25}\,\delta^\top P\delta .
\]
The full expectation of $\delta^\top P\delta=\av{n^2\one_D}$ is at least $1-\frac92d_0-\frac34\varrho$ by the preceding estimates, and outcomes that are not accepted contribute at most $\E[Z;\text{not accepted}]\le\frac3{128}+\tau_D$ to it. Hence
\[
 \E|u|^2\le\frac32-\frac{16}{25}\Big(1-\frac92d_0-\frac34\varrho-\frac3{128}-\tau_D\Big)=\frac78+\frac{16}{25}\Big(\frac92d_0+\frac34\varrho+\tau_D\Big).
\]
Since $|u|\le|\delta|$, its $L^2$ norm is at most $\sqrt{3/2}<\frac54$, and passing from $u$ to $w-r$ adds at most $\frac52E_*+E_*^2$. With \eqref{eq:double-shell} this proves \eqref{eq:Ubound}.
\end{proof}

\begin{proof}[Proof of Theorem~\ref{thm:menu}]
Assume \eqref{eq:hyp}. The inequalities used in Lemma~\ref{lem:rare} hold: $\theta^{96}=5^{96}\cdot10^{-2304}>1.2\cdot10^{-2237}$, so that $11\,\varepsilon=1.1\cdot10^{-2363}<10^{-126}\theta^{96}$, and $5\sqrt\varepsilon=5\cdot10^{-1182}<\theta^{16}/(392\cdot10^{17})$. Lemma~\ref{lem:arrival} gives $\Theta\le2\cdot10^{-45}$ and $E_0\le2\Theta/\varrho+6\varrho<2.1\cdot10^{-12}$. Since $\sqrt{10\varrho}<\frac32\cdot10^{-16}$, the bound \eqref{eq:gainzero} is $\mathcal J<3.5\cdot10^{-8}$, and \eqref{eq:cancel} gives
\[
 E_1<10^{-12}+3.4\cdot10^{-6}+2.8\cdot10^{-5}+2.03\cdot10^{-5}<5.2\cdot10^{-5}<d_0^2\qquad\text{with }d_0=0.0073 .
\]
With $\sqrt\varrho<10^{-16}$ and $\sqrt\tau<10^{-1181}$, \eqref{eq:Ubound} gives $E_*<0.0366$ and $\E|w-r|^2<0.875+0.0212+0.0915+0.0014<0.99$ for (S4). This contradicts \eqref{eq:lower}. The two inequalities used in Lemma~\ref{lem:rare}, the bounds on $E_0$, $\mathcal J$, $E_1$ and $E_*$ and the final bound $0.99$ are verified in exact arithmetic by the accompanying script \texttt{verify\_constants.py}.
\end{proof}

The size of $J$ comes from Lemma~\ref{lem:rare}: the threshold $\theta$ must be so small that $\Theta/(\varrho\gamma^2)$ is small, while $\varrho$ must be so small that $\cutoff\sqrt\varrho$ is small compared with $\gamma$, and the powers $\theta^{96}$ and $\theta^{16}$ enter through the smoothing in Lemma~\ref{lem:conc} and the cubes in Lemma~\ref{lem:three}. We did not try to improve these~steps.

\subsection{From the game to datasets}\label{app:transfer}

\begin{theorem}\label{thm:data}
For all $d\ge1$, $\kappa\ge1$ and $\zeta>0$ there is $n_0(d,\kappa,\zeta)$ such that for every $n\ge n_0$ and every dataset with one label per row, $L^*_D>0$ and $\kappa_D\le\kappa$ there is a selection with nonsingular design and $n\,\rho_D\le(c_*+\zeta)\,d\kappa$. In particular $\overline\alpha^{(1)}_{d,\kappa}\le c_*$.
\end{theorem}

\begin{proof}
Let $K_0(A)=\lceil J(1+A)\rceil$ with $J$ from Theorem~\ref{thm:U}. Thus the game admits cost at most $c_*C$ with densities in $[\frac12,5]$ and at most $K_0(A)$ counts.

Fix $\eta>0$ and set $R=d\kappa$ and
$m_0=1+d+d(d+1)/2$.
Whiten the features, subtract the full fit, and normalize the residuals
by $\sqrt{L^*}$, obtaining scalar residuals $e_i$ with
\[
 \frac1N\sum_ix_ix_i^\top=I,\quad |x_i|^2\le R,\quad\frac1N\sum_ix_ie_i=0,\quad\frac1N\sum_ie_i^2=1.
\]
The excess ratio of a fit $u$ to the residuals is exactly
$|u|^2$. Minimize $\sum p_i e_i^2$ subject to
$p_i\ge0$, $\sum p_i=1$, $\sum p_ix_ix_i^\top=I$, and
$\sum p_ix_ie_i=0$. This is a nonempty compact linear program.
It has $m_0$ equality constraints, so an optimal basic solution has at most
$m_0$ positive weights, and $\sum p_i e_i^2\le1$ because the uniform weights
are feasible; zero weights are discarded.
These auxiliary weights do not change the loss being evaluated.

Call a supported row rare if $|e_i|>\eta\sqrt n$.
The total probability $P$ of the rare rows is at most $\eta^{-2}/n$.
On rare rows set $a_i=np_i$, $g_i=x_ie_i/\sqrt n$, and
$r=\sum a_i g_i$. Then $A\le A_0:=\eta^{-2}$,
$C\le R$, and $|r|\le\sqrt{A_0R}$.
Put
\[
 T=\sum_{\mathrm{rare}}p_ix_ix_i^\top,\quad B=\frac{I-T}{1-P},\quad
 \nu_i=\frac{p_i}{1-P}\ (i\text{ frequent}),\quad x_i'=B^{-1/2}x_i.
\]
For all sufficiently large $n$, $B$ is positive definite and
$B^{\pm1/2}=I+O(n^{-1})$, uniformly over datasets, because
$\|T\|\le RP$.
Set $g_i'=B^{-1/2}g_i$, $r'=B^{-1/2}r$, and
$\phi_i=(1-P)\sqrt n\,e_i+x_i'\cdot r'$ on frequent rows.
Then $\E_\nu x'x'^\top=I$ and $\E_\nu\phi x'=0$, using
$\sum_{\mathrm{frequent}}p_ix_ie_i=-r/\sqrt n$.
Apply Theorem~\ref{thm:U} with $C'=\sum a_i|g_i'|^2$;
write its density as $1+h_i$ and its count as $K\le K_*:=K_0(A_0)$.
With $z=w-r'$, $v'=\sum k_i g_i'$,
$G'=\E_\nu(1+h_i)x_i'x_i'^\top$, and
$F=\E_\nu(1+h_i)\sqrt n\,e_ix_i'$, we have
\begin{equation}\label{eq:transferfit}
 \begin{gathered}
 |z|\le\sqrt{c_*C'},\qquad C'\le(1+O(n^{-1}))R,\qquad G'\succeq I/2,\\
 G'z=v'+(1-P)F,\qquad|F|\le5\sqrt{n/(1-P)}.
 \end{gathered}
\end{equation}
The last bound is Bessel's inequality and
$\E_\nu e^2\le1/(1-P)$.

For $n\ge K_*$, round $(n-K)\nu_i(1+h_i)$ to nonnegative integers
of total $n-K$, each with error at most one, by taking floors and
adding one to the required fractional entries. Include the rare counts.
Let $M_n,b_n$ be the resulting unnormalized design and residual
gradient, and put $\alpha=1-K/n$. Direct summation gives
\[
 \mathcal M:=B^{-1/2}(M_n/n)B^{-1/2}=\alpha G'+E_M,
 \quad\|E_M\|\le\|B^{-1}\|R(K_*+m_0)/n,
\]
\[
 \tilde b:=B^{-1/2}b_n/\sqrt n=v'+\alpha F+e_R,
 \quad |e_R|\le m_0\|B^{-1/2}\|\sqrt R\,\eta.
\]
Thus $\mathcal M\succeq I/3$ eventually. Subtracting
\eqref{eq:transferfit} gives the exact identity
\[
 \tilde b-\mathcal Mz=\frac KnG'z+(P-K/n)F+e_R-E_Mz.
\]
For fixed $d,\kappa,\eta$, all terms except $e_R$ are uniformly
$o(1)$: $\|G'\|\le5$, $K\le K_*$, $P\le A_0/n$, and
$|F|=O(\sqrt n)$. The actual residual fit satisfies
$\sqrt n\,u=B^{-1/2}\mathcal M^{-1}\tilde b$; hence
\[
 n|u|^2\le\left(\sqrt{c_*R}+3m_0\sqrt R\,\eta+o_n(1)\right)^2,
\]
uniformly over datasets. The selected design is positive definite. Choosing $\eta$ small and then $n$ large
proves the theorem.
\end{proof}

\section{Efficient selection with one label}\label{app:algo}

We use the notation of Appendix~\ref{app:one}.

\medskip
\begin{proof}[Proof of Theorem~\ref{thm:algo}]
Put $q=\max\{2,\lceil\log_2(n+1)\rceil\}$ and $\eta=2^{-\lceil q/16\rceil}$. If $n<d$, return any selection. Otherwise compute the full fit and, with one oracle call, an optimal basic solution $p$ of the linear program in the proof of Theorem~\ref{thm:data}, discard the rows with $p_i=0$, and call a remaining row rare if $|e_i|>\eta\sqrt n$. With $P$ and $T$ as there, if $P=1$ or $I-T$ is not positive definite, which does not happen for large $n$ because $P=O(n^{-7/8})$ and $\|T\|\le d\kappa P$, return a spanning set of rows repeated to size $n$; otherwise form the instance of the game as there. If the energy of the instance vanishes, use the single candidate $k=0$, $h=0$ in the following. Otherwise, for each of the strategies (S1) to (S4) make $H_\epsilon=4q\lceil\log_2(4/\epsilon)\rceil$ independent trials, with the cap $\overline K=\min\{n,\lceil q(1+A)\rceil\}$ in place of $\lceil J(1+A)\rceil$. For every trial with rare total $K$ round the frequent targets $(n-K)\nu_i(1+h_i)$ as there. Return the selection with the smallest excess loss on the full dataset among all candidates with nonsingular design, or a spanning set of rows repeated to size $n$ if there is none.

Let $n$ be so large that $q\ge J$ and $\lceil q(1+A)\rceil\le n$ for all $A\le\eta^{-2}$. Theorem~\ref{thm:menu} holds with any cap that is at least $\lceil J(1+A)\rceil$, since its proof uses only $\Prob(\sum_ik_i\le\overline K)\ge1-J^{-1}$. By Markov's inequality one of the strategies has cost at most $c\,C'$, where $C'$ is the energy of the instance as in the proof of Theorem~\ref{thm:data} and $c=1-1/(2J)$, with probability at least $1/(2J-1)$ in each trial, so all its $H_\epsilon$ trials fail with probability at most $\epsilon/4$. For a successful outcome the proof of Theorem~\ref{thm:data} applies with $A\le\eta^{-2}=O(n^{1/8})$, $P\le A/n$ and $K\le\overline K=O(q\,n^{1/8})$: now $\|T\|\le RP=O(n^{-7/8})$, hence $B^{\pm1/2}=I+O(n^{-7/8})$ and $C'\le(1+O(n^{-7/8}))R$, and $|z|\le\sqrt{cC'}$ replaces $|z|\le\sqrt{c_*C'}$. The error terms are $|e_R|=O(n^{-1/16})$, $\|E_M\|=O(q\,n^{-7/8})$, $|\frac KnG'z|=O(q\,n^{-7/8})$ and $|(P-K/n)F|=O(q\,n^{-3/8})$, uniformly over datasets, and $\mathcal M\succeq I/3$ for large $n$. Hence $n\,\rho\le(c+O_{d,\kappa}(n^{-1/16}))\,d\kappa$.

\emph{Sampling and computation.} For Poisson counts conditioned on $\sum_ik_i\le\overline K$, draw the total $K\in\{0,\dots,\overline K\}$ with weights $A^K/K!$ and allocate its arrivals independently with probabilities $a_i/A$. For the doubled law, draw totals $(U,V)$ with $U+2V\le\overline K$ and weights $\frac{(A/2)^U}{U!}\frac{(A/4)^V}{V!}$, allocate the $U$ single and the $V$ double arrivals separately with probabilities $a_i/A$, and put $k_i=N_i+2M_i$. The weights are computed by recursion; if $A=0$, all counts vanish. Let $\mathcal L$ bound the total number of categorical draws and $\mathcal S$ the number of outcomes of any of them; both are polynomial in $N$, $d$, $n$ and $\log(1/\epsilon)$, since $\overline K=O(q\,n^{1/8})$. Perform each draw by comparing a uniform dyadic number of $\lceil\log_2(4\mathcal L\mathcal S/\epsilon)\rceil$ bits with the cumulative probabilities. The error in total variation is at most $\mathcal S2^{-\lceil\log_2(4\mathcal L\mathcal S/\epsilon)\rceil}$ per draw, and a coupling of the successive draws bounds the total error probability by $\epsilon/4$, whatever the size of the smallest probability; with the trials, the failure probability is at most $\epsilon/2$. Integer parts of numbers in $[0,n]$ are found by bisection with $O(\log n)$ comparisons, and $q$, $\lceil\log_2(4/\epsilon)\rceil$ and the number of bits by repeated doubling. No square roots are needed. With raw rows and residuals $\hat e_i=\sqrt{L^*}\,e_i$, the linear program reads: minimize $\sum_ip_i\hat e_i^2$ over $p\ge0$ with $\sum_ip_i=1$, $\sum_ip_ix_ix_i^\top=\Sigma_D$ and $\sum_ip_ix_i\hat e_i=0$; it has the same feasible set and the same optimal basic solutions as the program in the proof of Theorem~\ref{thm:data}. The condition $I-T\succ0$ reads $\Sigma_D-\sum_{\mathrm{rare}}p_ix_ix_i^\top\succ0$, and the energy of the instance vanishes if and only if $x_i=0$ for every rare row. With $B_0=(\Sigma_D-\sum_{\mathrm{rare}}p_ix_ix_i^\top)/(1-P)$ and the common unit $1/\sqrt{nL^*}$, the projection of the $i$th rare gradient on the type $x$ is $x^\top B_0^{-1}x_i\hat e_i$, $s_x=-(1-P)\,n\,\hat e_x$, and $Z$ and $\sigma_x^2$ are rational in these numbers. The densities depend on ratios only, and the tests $|e_i|>\eta\sqrt n$ and $|\langle x,g_i\rangle|>\sigma_x/\sqrt\varrho$ are made after squaring. (S2) sorts at most $m_0$ values to find the weighted median, and (S3) and (S4) sum over pairs of supported rows. There are at most $4H_\epsilon$ candidates; after sampling, each is processed with a number of operations that is polynomial in $N$, $d$ and $n$.
\end{proof}

\section{Proof of Theorem~\ref{thm:barrier}}\label{app:barrier}

Proposition~\ref{prop:barrier} below bounds a family of datasets; parts (i) and (ii) choose its parameters, and Section~\ref{app:limits} bounds what lost poison and cover coordinates can contribute.

Throughout this appendix the labels are scalar, and $m$ denotes a total count of rare rows, not a number of labels. Write $(x)_+=\max\{x,0\}$, and $\langle\cdot,\cdot\rangle$ for the inner product.

\subsection{The construction}

Fix integers $M\ge3$, $1\le N<M-1$, $h\ge1$ and $T\ge1$, a rational $a>0$, a real $\beta>0$, and nonempty sets $S_1,\dots,S_T\subseteq[M]$ of sizes $\sigma_t$ with weights $\gamma_t>0$. Put
\[
 d=h+M+T,\qquad A=Ma,\qquad u=\frac1{a(M-1-N)} .
\]
There are $h$ \emph{heavy} coordinates $z_j$, $M$ \emph{poison} coordinates $b_l$ and $T$ \emph{cover} coordinates $s_t$. Let $r_{z_j}=h^{-1/2}$, $r_{b_l}=\sqrt\beta$, $r_{s_t}=\sqrt{\gamma_t}$, and for $i\in[M]$
\[
 g_{i,z_j}=\frac{h^{-1/2}}A,\qquad
 g_{i,b_l}=\begin{cases}-\sqrt\beta\,uN,&i=l,\\ \sqrt\beta\,u,&i\ne l,\end{cases}\qquad
 g_{i,s_t}=\frac{\sqrt{\gamma_t}}{\sigma_ta}\,\one_{i\in S_t}.
\]
Then $a\sum_ig_i=r$, $\|g_i\|\ge1/A$, and
\begin{equation}\label{eq:barC}
 C:=a\sum_i\|g_i\|^2=\frac1A+\frac{\beta M(N^2+M-1)}{a(M-1-N)^2}+\sum_t\frac{\gamma_t}{\sigma_ta},
\end{equation}
which does not depend on $h$. For $Q\subseteq[M]$ let $H(Q)=\beta|Q|+\sum_{t:S_t\cap Q=\varnothing}\gamma_t$, and choose $B_0$ with
\begin{equation}\label{eq:barB}
 0<B_0\le\min_{|Q|\le N}H(Q).
\end{equation}
In particular $B_0\le H(\varnothing)\le\|r\|^2$. For $\kappa>1$ the frequent rows will be multiples of the unit vectors, with limiting masses $\nu_{b_l}=\nu_{s_t}=1/(\kappa d)$ and $\nu_{z_j}=\nu_H/h$, where $\nu_H=1-(M+T)/(\kappa d)$.

\begin{proposition}\label{prop:barrier}
With $B_0$ as in \eqref{eq:barB}, for every real $\kappa>1$,
\[
 \underline\alpha^{(1)}_{d,\kappa}\ \ge\ \frac{X(\kappa)}C,\qquad X(\kappa)=\min\Big\{B_0,\ \Big(\Big(\frac{\nu_H(N+1)}A-1\Big)_+\Big)^2\Big\}.
\]
\end{proposition}

\begin{proof}
\emph{The datasets.} Let $\kappa$ be rational first. For large $n$ put $\alpha_n=1-A/n$ and $\pi_n=1/(\kappa d\alpha_n)$. Let $\nu^{(n)}_{b_l}=\nu^{(n)}_{s_t}=\pi_n$ and $\nu^{(n)}_{z_j}=\nu_H^{(n)}/h$ with $\nu_H^{(n)}=1-(M+T)\pi_n$. These masses are rational, sum to one, converge to the masses above, and are at least $\pi_n$ for large $n$, because $\nu^{(n)}_{z_j}-\pi_n={(1-d\pi_n)/h}>0$ if $n>\kappa A/(\kappa-1)$. Let $\epsilon_i=\|g_i\|/\sqrt{\kappa d\alpha_n}$ and $\xi_i=g_i/\epsilon_i$. For $w_0\in\R^d$ the dataset has the rare rows $(\xi_i,\ \langle\xi_i,w_0\rangle+\sqrt n\,\epsilon_i)$ of mass $a/n$ and the frequent rows $\big(\be_j/\sqrt{\nu^{(n)}_j},\ w_{0,j}/\sqrt{\nu^{(n)}_j}-r_j/(\alpha_n\sqrt{n\nu^{(n)}_j})\big)$ of mass $\alpha_n\nu^{(n)}_j$. The identity $a\sum g_i=r$ shows that the full fit is $w_0$; moreover $\Sigma=\alpha_nI+\frac an\sum\xi_i\xi_i^\top\succeq\alpha_nI$, and the optimal loss is $L^*_n=C/(\kappa d\alpha_n)+\|r\|^2/(n-A)\to C/(\kappa d)$. Every row has squared norm at most $\kappa d\alpha_n$, so the normalized leverage is at most $\kappa$.

\emph{Singular designs.} For fixed $n$ there are finitely many selections, and their designs do not depend on $w_0$. Choose $\bar w$ outside the orthogonal complements of the null spaces of the singular designs and $w_0=\omega\bar w$. The minimum-norm fit of a selection with null space $K$ is orthogonal to $K$, so its excess loss is at least $\alpha_n\omega^2\|\operatorname{pr}_K\bar w\|^2$, where $\operatorname{pr}_K$ is the orthogonal projection onto $K$. For $\omega$ large, every such selection has excess loss at least $\alpha_nB_0/n$. Neither $L^*_n$ nor the following bounds depend on $w_0$.

\emph{Nonsingular designs.} Let a selection have rare counts $k_i$ and frequent counts $c_j$; put $m=\sum k_i$, $Q=\{i:k_i>0\}$, $\tau=\sqrt n(\hat w-w_0)$, $\zeta_i=\langle\xi_i,\tau\rangle/(n\epsilon_i)$, $\tilde v_j=\sum_ik_ig_{i,j}(1-\zeta_i)$ and $e_n=1/\alpha_n$. The normal equations are
\begin{equation}\label{eq:barnormal}
 n\,\nu^{(n)}_j\,\tilde v_j=c_j\,(\tau_j+e_nr_j)\qquad\text{for all }j,
\end{equation}
and the excess loss is at least $\alpha_n\|\tau\|^2/n$. If $\|\tau\|^2>2\|r\|^2$, then $\|\tau\|^2\ge B_0$. Otherwise $|\zeta_i|\le\kappa d\,\|\tau\|/(n\|g_i\|)\le\bar\zeta_n:=2\kappa dA\|r\|/n$, uniformly over all selections, and we take $n$ so large that $\bar\zeta_n<1/(4N)$.

If $m=0$, all $c_j\ge1$ and \eqref{eq:barnormal} gives $\tau=-e_nr$, so $\|\tau\|^2\ge B_0$. Let $1\le m\le N$ and $l\in Q$. Then
\[
 \frac{\tilde v_{b_l}}{\sqrt\beta\,u}\le m-(N+1)k_l+\bar\zeta_n\big(m+(N-1)k_l\big)\le-k_l(1-2N\bar\zeta_n)<0,
\]
so \eqref{eq:barnormal} forces $c_{b_l}\ge1$ and $\tau_{b_l}<-e_n\sqrt\beta$. If $S_t\cap Q=\varnothing$, all selected rare features vanish in the coordinate $s_t$, hence $c_{s_t}\ge1$ and $\tau_{s_t}=-e_n\sqrt{\gamma_t}$. Therefore $\|\tau\|^2\ge H(Q)\ge B_0$ if $m\le N$.

Let $m\ge N+1$ and $y_n=(1-\bar\zeta_n)(N+1)/A$. In the heavy coordinates $\tilde v_{z_j}\ge h^{-1/2}y_n>0$, so $c_{z_j}\ge1$, and \eqref{eq:barnormal} with $p_j=c_{z_j}/n$ gives $\sqrt h\,\tau_{z_j}+e_n\ge\nu_H^{(n)}y_n/(hp_j)$. Since $\sum_jp_j\le1$, we have $\sum_j1/p_j\ge h^2$ and therefore $\sqrt h\sum_j\tau_{z_j}\ge h\,(\nu_H^{(n)}y_n-e_n)$; by the Cauchy--Schwarz inequality $\|\tau\|^2\ge\big((\nu_H^{(n)}y_n-e_n)_+\big)^2$.

Altogether every selection has excess loss at least $\alpha_nX_n/n$ with
\[
 X_n=\min\big\{B_0,\big((\nu_H^{(n)}y_n-e_n)_+\big)^2\big\}\ \to\ X(\kappa).
\]
Hence $n\,A_{d,\kappa}(n)\ge\alpha_nX_n/L^*_n$, which proves the claim for rational $\kappa$. Since $X$ is continuous and $A_{d,\kappa}(n)$ is nondecreasing in $\kappa>1$, we may approximate a real $\kappa$ from below.
\end{proof}

\subsection{Part (i)}

For $c\le0$ the claim is trivial. Let $0<c<e^{-1/2}$, choose $\eta\in(0,\frac12)$ with $e^{-1/2}/(1+2\eta)>c$, and put $\alpha=\sqrt e/(1+2\eta)$ and $\nu^*=(1+\eta)/(1+2\eta)$. For an integer $v$ so large that $N+\sigma<M-1$ below, let $a=1/v$, $M=v^3$, $A=v^2$, $N=\lceil(1+2\eta)v^2\rceil$, $\sigma=\lceil\alpha v\rceil$, $\beta=\alpha\eta^2/(2A)$, $\lambda=\sigma/(M-N-\sigma)$ and $\gamma=e\beta/\lambda$. Let the covers be the $T=\binom M\sigma$ subsets of $[M]$ of size $\sigma$, each with weight $\gamma/T$. A set $Q$ with $q\le N$ elements misses the fraction
\[
 \prod_{0\le l<q}\Big(1-\frac{\sigma}{M-l}\Big)\ \ge\ \exp\Bigl(-\sum_{0\le l<q}\frac\sigma{M-l-\sigma}\Bigr)\ \ge\ e^{-\lambda q}
\]
of the covers, by $\log(1-x)\ge-x/(1-x)$. Hence $H(Q)\ge\beta q+\gamma e^{-\lambda q}\ge\min_x(\beta x+\gamma e^{-\lambda x})=2\beta/\lambda=:B_0$. If $\nu_H\ge\nu^*$, then $\nu_H(N+1)/A\ge1+\eta$, and Proposition~\ref{prop:barrier} gives $\underline\alpha^{(1)}_{d,\kappa}\ge R_v$ with $R_v=\min\{\eta^2,2\beta/\lambda\}/C$. As $v\to\infty$, $\lambda A\to\alpha$, so $2\beta/\lambda\to\eta^2$; in \eqref{eq:barC} the first term tends to $0$, and the second and the third both tend to $\frac12\alpha\eta^2(1+2\eta)^2$. Hence $R_v\to{1/(\alpha(1+2\eta)^2)}=e^{-1/2}/(1+2\eta)>c$. Fix $v$ with $R_v>c$; neither $v$ nor the covers depend on $\kappa$. For $d\ge d_1:=\lceil(M+T)/(1-\nu^*)\rceil$ take $h=d-M-T$; then $\nu_H\ge\nu^*$ and $\underline\alpha^{(1)}_{d,\kappa}\ge R_v>c$.

\subsection{Part (ii)}

\emph{Covers from a Steiner system.} Let $\mathbb K=\mathbb F_9\subset\mathbb E=\mathbb F_{81}$ with norm $\operatorname{Nm}(x)=x^{10}$. On the $82$ points $\mathbb E\cup\{\infty\}$ let the blocks be the affine $\mathbb K$-lines of $\mathbb E$ with $\infty$ adjoined and the circles ${\{x:\operatorname{Nm}(x-c)=\varrho\}}$, $c\in\mathbb E$, $\varrho\in\mathbb K\setminus\{0\}$. Every block has ten points. A triple containing $\infty$ lies on exactly one line. Three collinear points of $\mathbb E$ lie on their line and on no circle, since the norm equation restricted to a line is a quadratic with nonzero leading coefficient. For three noncollinear points, subtracting the circle equations gives two independent linear equations for the center, because the form $xy^9+x^9y$ is nondegenerate; the common value $\varrho$ is nonzero. So every triple lies in exactly one block: this is a Steiner system $S(3,10,82)$, with $738$ blocks, $90$ through each point and $10$ through each pair. By $\one_{v=0}\ge1-v+\binom v2-\binom v3$ for integers $v\ge0$, a set of $j$ points misses at least
\[
 738-90j+10\binom j2-\binom j3=598-45j+\frac{(j-6)(j-7)(20-j)}6
\]
blocks, and hence at least $598-45j$ blocks for every $j$: for $j\le20$ the last term is nonnegative, and for $j\ge14$ the number $598-45j$ is negative.

\emph{Parameters.} Take $270$ disjoint copies of the Steiner system, replace each point by $6$ rare rows, and let the covers be the $60$ rows over each block, all with weight $\gamma_0=3.215\cdot10^{-8}$. Let $\beta=45\gamma_0$, $a=\frac1{46}$ and $N=3098$. Then $M=82\cdot6\cdot270=132\,840$, $T=738\cdot270=199\,260$ and $A=\frac{66\,420}{23}$. If $Q$ touches $j_l$ points in the $l$th copy, then $\sum_lj_l\le|Q|$, so that
\[
 H(Q)\ \ge\ \sum_l\big(\beta j_l+\gamma_0(598-45j_l)\big)=598\cdot270\,\gamma_0=\frac{5\,190\,939}{10^9}=:B_0\qquad\text{for every }Q ,
\]
and by \eqref{eq:barC}
\[
 C=\frac{643\,994\,347\,078\,906\,254\,541}{62\,112\,762\,928\,890\,000\,000\,000},\qquad\frac{B_0}C>0.5006 .
\]
Let $d\ge332\,101$, $d\kappa\ge332\,100\,000$ and $h=d-332\,100$. Then $\nu_H=1-332\,100/(\kappa d)\ge\frac{999}{1000}$, so that $\nu_H(N+1)/A-1>0.07205$ and its square exceeds $B_0$, and Proposition~\ref{prop:barrier} proves part~(ii). The finite-field incidences and the rational comparisons are checked in exact arithmetic by the accompanying script \texttt{verify\_constants.py}.

\subsection{Limits of the construction}\label{app:limits}

In the construction a coordinate is lost if the selected gradient does not have the sign of the target there, and the total count is capped. We show that for rare rows with a common coefficient and for large $A$, some selection of total count between $1$ and $A$ loses covers of arbitrary sets and poisons with one exceptional entry and a common positive entry elsewhere whose squared targets sum to at most the fraction $e^{-1/2}+O(1/A)$ of the energy of the whole family. Let $M$ rare rows have the common coefficient $a\in(0,1]$ and let $A=Ma$. For $\ell\in\R^M$ put $\mu(\ell)=a\sum_i\ell_i$ and $E(\ell)=a\sum_i\ell_i^2$; these are the target and the energy of a coordinate in which the $i$th rare gradient has the entry $\ell_i$. Suppose $\mu(\ell)>0$ and say that $\ell$ is \emph{lost} under $k\in\Z_{\ge0}^M$ if $\sum_ik_i\ell_i\le0$. A \emph{cover} is $\ell=\one_S$ with $\varnothing\ne S\subseteq[M]$; a \emph{poison} is $\ell_i=1-bA\,\one_{i=l}$ with $l\in[M]$, $b>0$ and $ab<1$, so that $\mu(\ell)=A(1-ab)>0$. (In the construction, $b=(N+1)/A$.)

\begin{proposition}\label{prop:limits}
There are absolute constants $A_0$ and $K_0$ such that for every integer $M\ge1$ and every $a\in(0,1]$ with $A=Ma\ge A_0$ there is a random $k\in\Z_{\ge0}^M$ with $1\le\sum_ik_i\le A$ and $k_i\le8$ for which
\[
 \mu(\ell)^2\,\Prob(\ell\text{ is lost})\le\Big(e^{-1/2}+\frac{K_0}A\Big)E(\ell)
\]
for every cover and every poison. Consequently, for every finite family of positive multiples of covers and poisons with total energy $C$, some $k$ with $1\le\sum_ik_i\le A$ loses coordinates with total squared target at most $(e^{-1/2}+K_0/A)\,C$.
\end{proposition}

\begin{proof}
Let $v=e^{-1/2}$. A \emph{profile} is a sequence of tails $\rho_1\ge\dots\ge\rho_8\ge\rho_9=0$; put $x_j=\rho_j-\rho_{j+1}$ and $\tau=\sum_j\rho_j=\sum_jjx_j$. We use two profiles, $\rho_j=v\alpha_j$ with probability $\frac35$ and $\rho_j=v\beta_j$ with probability $\frac25$, where
\begin{center}\small
\begin{tabular}{@{}lcccccccc@{}}
\toprule
$j$ & $1$&$2$&$3$&$4$&$5$&$6$&$7$&$8$\\
\midrule
$\alpha_j$ & $1$ & $\sqrt e-1.34$ & $.15$ & $.08$ & $.05$ & $.03$ & $.02$ & $.01$\\
$\beta_j$ & $1$ & $.14$ & $.04$ & $.02$ & $.02$ & $.02$ & $.02$ & $.02$\\
\bottomrule
\end{tabular}
\end{center}
Their values of $\tau$ are $1$ and $\theta=1.28\,v$. Given the profile, assign the multiplicity $j$ to $\lfloor Ax_j\rfloor$ rows, for $j=1,\dots,8$, where these disjoint sets of rows are placed uniformly at random, and the multiplicity zero to all other rows. This is possible because $\sum_jAx_j=A\rho_1\le M$, and
\begin{equation}\label{eq:limcount}
 A\tau-36\le\sum_ik_i\le A\tau\le A,\qquad |\{i:k_i>0\}|\ge Av-8 .
\end{equation}

\emph{Covers.} Let $|S|=\sigma$ and $\eta=a\sigma$, so $\mu^2/E=\eta$. If $R$ rows have positive multiplicity, $S$ is missed with probability at most $(1-R/M)^\sigma\le e^{-\eta R/A}$, and $\eta e^{-\eta R/A}\le A/(eR)\le1/(e(v-8/A))$, which is $v+O(1/A)$ because $1/(ev)=v$.

\emph{Poisons.} Since $\sum_ik_i\ell_i=\sum_ik_i-bAk_l$ and $\sum_ik_i\ge1$, a poison is lost only if $k_l=j\ge1$ and $jbA\ge\sum_ik_i$. By \eqref{eq:limcount} this implies $jb'\ge\tau$ with $b'=b+36/A$. Given the profile, $k_l=j$ has probability at most $Ax_j/M=ax_j$. Hence
\[
 \Prob(\ell\text{ is lost})\le a\,\Phi(b'),\qquad
 \Phi(b)=v\Big(\tfrac35\,\alpha_{\lceil1/b\rceil}+\tfrac25\,\beta_{\lceil\theta/b\rceil}\Big),
\]
with $\alpha_j=\beta_j=0$ for $j>8$. We claim that $\Phi(b)\le vb^2$ for all $b>0$. As $\Phi$ is a nondecreasing step function that changes only at the points $1/j$ and $\theta/j$, $1\le j\le8$, and is right-continuous, it suffices to check these points. With $\theta^2=1.28^2/e>0.6027$ and $\sqrt e-1.34<0.3088$ the values are as follows (upper bounds for $\Phi/v$ and lower bounds for $b^2$, exact or rounded in the safe direction).
\begin{center}\small\renewcommand{\arraystretch}{1.25}
\begin{tabular}{@{}lcccccccc@{}}
\toprule
$b$ & $1$ & $\theta$ & $\frac12$ & $\frac\theta2$ & $\frac13$ & $\frac\theta3$ & $\frac14$ & $\frac15$\\
$(\lceil\frac1b\rceil,\lceil\frac\theta b\rceil)$ & $(1,1)$ & $(2,1)$ & $(2,2)$ & $(3,2)$ & $(3,3)$ & $(4,3)$ & $(4,4)$ & $(5,4)$\\
$\Phi(b)/v$ & $1$ & $.5853$ & $.2413$ & $.146$ & $.106$ & $.064$ & $.056$ & $.038$\\
$b^2$ & $1$ & $.6027$ & $.25$ & $.1506$ & $.1111$ & $.0669$ & $.0625$ & $.04$\\
\midrule
$b$ & $\frac\theta4$ & $\frac16$ & $\frac\theta5$ & $\frac17$ & $\frac\theta6$ & $\frac18$ & $\frac\theta7$ & $\frac\theta8$\\
$(\lceil\frac1b\rceil,\lceil\frac\theta b\rceil)$ & $(6,4)$ & $(6,5)$ & $(7,5)$ & $(7,6)$ & $(8,6)$ & $(8,7)$ & $(10,7)$ & $(11,8)$\\
$\Phi(b)/v$ & $.026$ & $.026$ & $.020$ & $.020$ & $.014$ & $.014$ & $.008$ & $.008$\\
$b^2$ & $.0376$ & $.0277$ & $.0241$ & $.0204$ & $.0167$ & $.0156$ & $.0123$ & $.0094$\\
\bottomrule
\end{tabular}
\end{center}
For $b\ge1$ we have $\Phi(b)=v\le vb^2$, and $\Phi(b)=0$ for $b<\theta/8$. This proves the claim. Moreover, with $0<1-ab<1$ because $\mu>0$,
\[
 E=A\big((1-ab)^2+ab^2(A-a)\big),\qquad
 \frac{\mu^2}E=\frac{A(1-ab)^2}{(1-ab)^2+ab^2(A-a)}\le\frac1{ab^2(1-1/A)} .
\]
If the poison is lost with positive probability, then $b'\ge\theta/8$, hence $b'/b=1+36/(Ab)=1+O(1/A)$ and
\[
 \frac{\mu^2}E\,\Prob(\ell\text{ is lost})\le\frac{v}{1-1/A}\Big(\frac{b'}b\Big)^2=v+O(1/A).
\]
The last assertion follows by summing over the family and taking an outcome that is no worse than the mean.
\end{proof}

The constant $e^{-1/2}$ cannot be lowered uniformly in $a\in(0,1]$ as $A\to\infty$. In the construction of part~(i), with $\beta$, $\lambda$, $\eta$, $N$ and $C$ as there and $a=A^{-1/2}\to0$, every selection of total count at most $A$ has at most $A\le N$ distinct rows and loses poison and cover coordinates with total squared target at least $2\beta/\lambda$, and the energy of all poison and cover coordinates is $C-1/A$. Dividing the guaranteed lower bound $2\beta/\lambda$ by $C-1/A$ gives a quantity that tends to $e^{-1/2}/(1+2\eta)$, with $\eta>0$ arbitrary.

\section{Proof of Theorem~\ref{thm:group}}\label{app:group}

This appendix proves parts (ii) and (iii) of Theorem~\ref{thm:group}; part (i) is proved in Section~\ref{sec:group}. For $d=1$, part (ii) is the one-dimensional case of mean selection, whose value is $1/(2n-1)$ \citep{hanneke2025erm}. It also follows from Lemma~\ref{lem:twopoint} with $b=n$: if some label equals the mean, $n$ copies of it have error zero; otherwise $k=0$ if $np\le\frac12$ and an integer $k$ nearest to $np$ otherwise give $(k-np)^2/(n^2p(1-p))\le1/(2n-1)$. A group in which a fraction $1/(2n)$ of the labels equals $z^+$ and the others equal $z^-\ne z^+$ attains this value: a selection with $k$ copies of $z^+$ has ratio $(2k-1)^2/(2n-1)$.

In a grouped dataset let $\nu_j$ be the mass of group $j$, and $\mu_j$ and
$\sigma_j^2$ the mean and variance of its labels. Then $w^*=(\mu_j)_j$ and
$L^*=\sum_j\nu_j\sigma_j^2$. A selection gives group $j$ a multiset of
$b_j\ge0$ labels with $\sum_jb_j=n$; the minimum-norm fit has $\hat w_j$
equal to their mean if $b_j\ge1$ and $\hat w_j=0$ if $b_j=0$, and the
excess loss is $\sum_j\nu_j(\hat w_j-\mu_j)^2$. We write
$\vartheta_j=\nu_j\sigma_j^2/L^*$, so that $\sum_j\vartheta_j=1$.

\subsection{The upper bound (ii)}

\begin{lemma}[Two labels]\label{lem:twopoint}
Consider a group with label mean $\mu$ and variance $\sigma^2$. Either
some label of the group equals $\mu$, or there are two labels $z^+\ne z^-$
of the group and $p\in(0,\frac12]$ such that for all integers $b\ge1$ and
$0\le k\le b$ the mean $\hat w$ of $k$ copies of $z^+$ and $b-k$ copies of
$z^-$ satisfies
\[
 (\hat w-\mu)^2\ \le\ \sigma^2\,\frac{(k-bp)^2}{b^2\,p\,(1-p)} .
\]
\end{lemma}

\begin{proof}
Minimize $\sum_iw_i(z_i-\mu)^2$ over probability vectors $w$ on the rows of
the group, with labels $z_i$, subject to $\sum_iw_iz_i=\mu$. The feasible
set is a nonempty polytope (the uniform vector is feasible, with value
$\sigma^2$), so the minimum is attained at a vertex. Since there are two
equality constraints, a vertex has at most two nonzero entries, and the
columns $(1,z_i)$ of its support are linearly independent, so the labels in
its support are distinct. If the support is one row, its label equals
$\mu$. Otherwise the support consists of two rows with labels $z^+\ne z^-$,
where $z^+$ denotes one of weight $p\le\frac12$. Then $pz^++(1-p)z^-=\mu$
and $p(1-p)(z^+-z^-)^2\le\sigma^2$, and $\hat w-\mu=(k/b-p)(z^+-z^-)$.
\end{proof}

\begin{lemma}[Allocation]\label{lem:alloc}
Let $\vartheta_1,\dots,\vartheta_q\ge0$ with $\sum_j\vartheta_j\le1$, let
$c_1,\dots,c_q>0$, $w_j=\vartheta_j/c_j$ and $B>0$. For a vector of levels
$o\in\{0,\frac34,1\}^q$ let its budget be $\sum_jo_jc_j$ and its cost
$\sum_j\gamma(o_j)w_j$, where $\gamma(0)=1$, $\gamma(\frac34)=\frac19$ and
$\gamma(1)=0$. There is a level vector with budget at most $B$ and cost at
most
\[
 \frac1B\Big(\frac{27\,q}{128}+\frac{13}{24}\Big) .
\]
\end{lemma}

\begin{proof}
Each group has two \emph{steps}: from level $0$ to $\frac34$, which
increases the budget by $\frac34c_j$ and decreases the cost by $\frac89w_j$,
and from $\frac34$ to $1$, which increases the budget by $\frac14c_j$ and
decreases the cost by $\frac19w_j$. Their ratios of cost decrease to budget
increase are $\frac{32}{27}w_j/c_j$ and $\frac49w_j/c_j$. Order the $2q$
steps by decreasing ratio, the first step of a group before its second.
Starting from all levels zero, perform the steps in this order as long as
the budget stays at most $B$, and stop at the first step that does not
fit. If all steps are performed, all levels equal one and the cost is
zero. Otherwise let $j_0$ be the group of the blocked step, $\Delta$ its
budget increase and $\omega$ its ratio, and let $U$ be the budget of the
level vector $o$ obtained, so that $U\le B<U+\Delta$.

With $h_j(s)=\gamma(s)w_j+\omega\,s\,c_j$ for $s\in\{0,\frac34,1\}$ we have
$h_j(\frac34)-h_j(0)=\frac34c_j(\omega-\frac{32}{27}w_j/c_j)$ and
$h_j(1)-h_j(\frac34)=\frac14c_j(\omega-\frac49w_j/c_j)$. A performed step has
ratio at least $\omega$, a step that is not performed has ratio at most
$\omega$, and the second ratio of a group is at most its first. Hence $o_j$
minimizes $h_j$ over the three levels, for every $j$. With
$g(x)=\min\{1,\frac19+\frac34x,x\}$ this minimum equals
$w_jg(\omega c_j/w_j)$ if $w_j>0$ and zero otherwise. Let
$\kappa_0=(27/32)^{1/2}$. We claim that $g(x)\le\kappa_0\sqrt x$ for
$x\ge0$. Indeed $x\le\kappa_0\sqrt x$ for $x\le\kappa_0^2$,
$1\le\kappa_0\sqrt x$ for $x\ge\kappa_0^{-2}$, and for
$\kappa_0\le\sqrt x\le\kappa_0^{-1}$
\[
 \frac19+\frac34x-\kappa_0\sqrt x=\frac34\Big(\sqrt x-\frac1{\kappa_0}\Big)\Big(\sqrt x-\frac{4\kappa_0}{27}\Big)\le0 .
\]
Hence the minimum is at most
$\kappa_0(\omega\,c_jw_j)^{1/2}=\kappa_0(\omega\vartheta_j)^{1/2}$, and with
$S=\sum_j\vartheta_j^{1/2}$ the cost of $o$ satisfies
\begin{equation}\label{eq:alloc}
 \sum_j\gamma(o_j)w_j=\sum_jh_j(o_j)-\omega U\ \le\ \kappa_0\sqrt{\omega}\,S-\omega U .
\end{equation}
The blocked step lowers the cost by $\omega\Delta$. For a first step $\omega\vartheta_{j_0}=\frac{32}{27}w_{j_0}^2$, so that $\omega\Delta=\frac89w_{j_0}=\frac89\kappa_0(\omega\vartheta_{j_0})^{1/2}$; for a second step $\omega\vartheta_{j_0}=\frac49w_{j_0}^2$, so that $\omega\Delta=\frac19w_{j_0}=\frac16(\omega\vartheta_{j_0})^{1/2}$, whose coefficient is smaller. Put $y=\vartheta_{j_0}^{1/2}$. By \eqref{eq:alloc} and $U>B-\Delta$ the cost is at most
\[
 \kappa_0\sqrt\omega\,S-\omega B+\omega\Delta\ \le\ \kappa_0\sqrt\omega\,\big(S+\tfrac89y\big)-\omega B\ \le\ \frac{\kappa_0^2}{4B}\big(S+\tfrac89y\big)^2=\frac{27}{128\,B}\big(S+\tfrac89y\big)^2 .
\]
Since $S\le y+((q-1)(1-y^2))^{1/2}$, the Cauchy--Schwarz inequality gives
\[
 \big(S+\tfrac89y\big)^2\ \le\ \Big(\tfrac{17}9y+\sqrt{q-1}\,(1-y^2)^{1/2}\Big)^2\ \le\ \tfrac{289}{81}+q-1=q+\tfrac{208}{81},
\]
and $\frac{27}{128}\cdot\frac{208}{81}=\frac{13}{24}$.
\end{proof}

\begin{proof}[Proof of the bound $27d/128+13/24$]
In this proof $o(1)$ denotes quantities that are bounded by null sequences depending on $n$ and $d$ only, and $n$ is large. Let a grouped
dataset with $L^*>0$ be given and apply Lemma~\ref{lem:twopoint} to every group. Let $E$ be the set
of groups in which some label equals the group mean; such a group
contributes nothing to the excess loss when it receives any positive number
of copies of that label. For the other groups let $z_j^\pm$ and $p_j$ be as
in the lemma. If such a group receives $k$ copies of $z_j^+$ and $b-k$
copies of $z_j^-$, its contribution to $n\,\rho$ is at most
$\vartheta_j\,\Psi_j(b,k)$ with
\[
 \Psi_j(b,k)=\frac{n\,(k-bp_j)^2}{b^2\,p_j\,(1-p_j)} .
\]
Let $F$ be the set of these groups with $np_j\ge n^{1/4}$ and $R$ the set of
those with $np_j<n^{1/4}$.

\emph{Groups in $F$.} If $b\ge n^{15/16}$ and $k$ is an integer nearest
to $bp_j$, then $0\le k\le b$ and
$\Psi_j(b,k)\le n/(4b^2p_j(1-p_j))\le n/(2b^2p_j)\le\frac12n^{-1/8}$.

\emph{Groups in $R$.} Here $p_j<n^{-3/4}\le\frac14$. Put $c_j=1/(np_j)$; then
$\Psi_j(b,k)=(k/t-1)^2/(c_j(1-p_j))$ with $t=bp_j$. We use three levels.
Level $0$: $k=0$ and $b=1$, with $\Psi_j=1/(c_j(1-p_j))$; the same value
holds for $k=0$ and every $b\ge1$. Level $\frac34$: $k=1$ and
$b=\lceil3/(4p_j)\rceil$; then $t\in[\frac34,1]$ and
$\Psi_j\le1/(9c_j(1-p_j))$. Level $1$: $k=1$ and $b=\lceil1/p_j\rceil$; then
$t\in[1,1+p_j)$ and $\Psi_j\le p_j^2/(c_j(1-p_j))\le2n^{-5/4}$. The number
of rows of group $j$ at level $o_j$ is at least $n\,o_jc_j$ and at most
$n\,o_jc_j+1$.

\emph{Budget.} Give one row to each group in $E$ and
$\lceil n^{15/16}\rceil$ rows to each group in $F$. This leaves
$n_R=n-o(n)$ rows for the groups in $R$. If $R$ is empty,
then $E\cup F$ is nonempty; add the remaining rows to one of its groups as
in case (a) below, so that $n\,\rho=o(1)$. Let now $R$ be
nonempty, and apply Lemma~\ref{lem:alloc} to the groups in $R$, with $q=|R|\le d$, the
numbers $\vartheta_j$ and $c_j$, and $B=(n_R-d)/n=1-o(1)$. The resulting
levels use at most $nB+d=n_R$ rows, the groups in $R$ contribute at most $27d/128+13/24+o(1)$
to $n\,\rho$, and the groups in $F$ contribute $o(1)$. It remains to
place the rows that are left over, say $r_0\ge0$ of them, without
increasing these bounds.

\emph{(a)} If $E\cup F$ is nonempty, add the $r_0$ rows to one of its
groups; the bounds for $E$ and $F$ hold for every larger number of rows
(in a group of $F$, with $k$ again an integer nearest to $bp_j$).

\emph{(b)} Otherwise, if some group in $R$ is at level $0$, add $r_0$
copies of $z_j^-$ to such a group $j$; this keeps~$k=0$.

\emph{(c)} Otherwise, if some group is at level $\frac34$, give it the $r_0$ rows and let $k$ be an integer nearest to $t=bp_j$, where $b$ is its new number of rows. Since $t\ge\frac34$, we have $|k-t|\le|1-t|\le t/3$ if $t\le\frac32$, and $|k-t|\le\frac12\le t/3$ otherwise, so that the bound $\Psi_j\le1/(9c_j(1-p_j))$ remains valid.

\emph{(d)} Otherwise $E=F=\varnothing$, $B=1-d/n$ and all groups are at level $1$. Put $\bar c=\sum_jc_j\le B$. The rows used are at most $n\bar c+d$, so $r_0\ge n(B-\bar c)$. Some group has $\vartheta_j/c_j\le1/\bar c$, because $\sum_j\vartheta_j\le1$; give it the $r_0$ rows and let $k$ be an integer nearest to $t=bp_j$, where $b$ is its new number of rows. Then $t\ge1+(B-\bar c)/c_j\ge B/\bar c$, because $c_j\le\bar c$, and its contribution is at most $\vartheta_j/(4c_jt^2(1-p_j))\le\bar c/(4B^2(1-p_j))\le1/(4B(1-p_j))$. The other groups contribute $o(1)$, so $n\,\rho\le\frac14+o(1)$.

In all cases $n\,\min_k\rho(k)\le27d/128+13/24+o(1)$.
\end{proof}

\subsection{The label-blind constant (iii)}

For the upper bound let $n\ge d$ and give every group $\lfloor n/d\rfloor$ rows drawn
independently and uniformly from the group, and the remaining rows to the
first group in the same way. The expected excess loss is at most
$\sum_j\nu_j\sigma_j^2/\lfloor n/d\rfloor=L^*/\lfloor n/d\rfloor$. For the
lower bound let the design have $d$ groups of $M$ rows each, $M\ge n+2$ even, and
let $Q$ be a law on selections. If $Q(b_j=0)>0$ for some $j$, let all
labels in group $j$ be $\mu\pm1$ with balanced signs and all other labels
zero; then $L^*=1/d$ and the expected excess loss is at least
$Q(b_j=0)\mu^2/d$, which is unbounded in $\mu$. Otherwise all $b_j\ge1$
almost surely, and some group $j$ has $\E_Qb_j\le n/d$. Give the rows of
this group uniformly random balanced signs and all other rows the label
zero. As in the proof of \eqref{eq:fam1}, the expected squared error of
the mean of a multiset of $b_j$ labels is at least $(1-n/(M-1))/b_j$, so by
Jensen's inequality the expected excess ratio is at least
$(1-n/(M-1))\,\E_Q[1/b_j]\ge(1-n/(M-1))\,d/n$. Letting $M\to\infty$ gives
$n\,B^{\mathrm{grp}}_d(n)\ge d$ for every $n\ge d$, and with the upper bound,
$\lim_nn\,B^{\mathrm{grp}}_d(n)=d$.

\end{document}